\documentclass{article}
\usepackage{etoolbox}

\newcommand{\venue}{arxiv}

\ifdefstring{\venue}{neurips}{%
  \usepackage[dblblindworkshop]{neurips_2026}%
  \workshoptitle{Transitioning from Pre-Training to Post-Training}%
  \newcommand{\venuebst}{plainnat}%
}{\ifboolexpr{test {\ifdefstring{\venue}{iclr}} or test {\ifdefstring{\venue}{arxiv}}}{%
  \usepackage{iclr2027_conference}%
  \ifdefstring{\venue}{arxiv}{%
    \iclrfinalcopy
    \renewcommand{\headrulewidth}{0pt}%
    \setlength{\textheight}{\dimexpr\paperheight-2in-\voffset-\topmargin-\headheight-\headsep-\maxdepth\relax}%
    \expandafter\patchcmd\csname @maketitle\endcsname
      {\lhead{Published as a conference paper at ICLR 2027}}{\lhead{}}{}%
      {\errmessage{Could not remove the ICLR banner for arxiv}}%
  }{}%
  \usepackage{environ}%
  \ificlrfinal
  \else
    \NewEnviron{ack}{}%
  \fi
  \newcommand{\venuebst}{iclr2027_conference}%
  \ifdefstring{\venue}{iclr}{%
  }{%
  }%
}{%
  \errmessage{Unknown venue `\venue': expected neurips, iclr, or arxiv}%
}}

\usepackage[utf8]{inputenc}
\usepackage[T1]{fontenc}
\usepackage{hyperref}
\usepackage{url}
\usepackage{booktabs}
\usepackage{multirow}
\usepackage{placeins}
\usepackage{tabularx}
\usepackage{amsfonts}
\usepackage{amsmath}
\usepackage{graphicx}
\usepackage{listings}
\usepackage{nicefrac}
\usepackage{microtype}
\usepackage{xcolor}
\usepackage[listings,breakable,skins]{tcolorbox}

\lstdefinestyle{promptstyle}{
  basicstyle=\ttfamily\fontsize{10}{11}\selectfont,
  breaklines=true,
  breakindent=0pt,
  columns=fullflexible,
  keepspaces=true,
  upquote=true,
  extendedchars=true,
  aboveskip=2pt, belowskip=0pt,
}

\newcommand{\promptheadfont}{\sffamily\bfseries\fontsize{9}{11}\selectfont}
\newcommand{\promptlabel}[1]{{\promptheadfont #1\par\smallskip}}
\newcommand{\promptpart}[1]{{\promptheadfont #1\par\smallskip}}

\newtcolorbox{promptbox}[2][]{
  breakable,
  bicolor,
  colback=white,
  colbacklower=black!5,
  colframe=black,
  coltitle=white, colbacktitle=black,
  fonttitle=\promptheadfont,
  title={#2},
  arc=3mm, boxrule=1pt,
  left=4mm, right=4mm, top=2mm, bottom=2mm,
  before upper={\promptlabel{SYSTEM PROMPT}},
  before lower={\promptlabel{USER PROMPT}},
  #1
}

\lstdefinestyle{figcode}{%
  basicstyle=\ttfamily\tiny,
  breaklines=true,
  breakautoindent=true,
  columns=fullflexible,
  keepspaces=true,
  frame=single,
  framesep=2.5pt,
  xleftmargin=3pt,
  xrightmargin=1pt,
  aboveskip=2pt,
  belowskip=2pt,
}

\newcommand{\papersource}{original}

\ifdefstring{\papersource}{original}{%
}{%
}

\newcommand{\method}{Don't Repeat Yourself Supervised Fine-Tuning}
\newcommand{\methodshort}{\texorpdfstring{\mbox{DRY-SFT}}{DRY-SFT}}

\newcommand{\passk}[1]{pass@#1}
\newcommand{\passkplus}[1]{pass@#1+}

\newcommand{\HEBaseCovOne}{0.757}
\newcommand{\HEBaseCovHundred}{0.851}
\newcommand{\HEBaseAst}{0.072}
\newcommand{\HECtrlAst}{0.043}
\newcommand{\HEPSFTAst}{0.053}
\newcommand{\HEDivCovOne}{0.735}
\newcommand{\HEDivCovHundred}{0.960}
\newcommand{\HEDivAst}{0.264}
\newcommand{\HEChainAst}{0.403}
\newcommand{\HEGainHundred}{10.8}
\newcommand{\HECtrlGainHundred}{0.7}
\newcommand{\HEPSFTGainHundred}{1.0}
\newcommand{\HEPSFTBestEpoch}{2}
\newcommand{\MBBaseCovOne}{0.665}
\newcommand{\MBBaseCovHundred}{0.749}
\newcommand{\MBBaseAst}{0.051}
\newcommand{\MBCtrlAst}{0.058}
\newcommand{\MBPSFTAst}{0.060}
\newcommand{\MBDivCovOne}{0.648}
\newcommand{\MBDivCovHundred}{0.874}
\newcommand{\MBDivAst}{0.286}
\newcommand{\MBChainAst}{0.453}
\newcommand{\MBGainHundred}{12.5}
\newcommand{\MBCtrlGainHundred}{1.8}
\newcommand{\MBPSFTGainHundred}{2.5}
\newcommand{\MBPSFTBestEpoch}{1}
\newcommand{\DSBaseCovOne}{0.357}
\newcommand{\DSBaseCovHundred}{0.507}
\newcommand{\DSBaseAst}{0.057}
\newcommand{\DSCtrlAst}{0.073}
\newcommand{\DSPSFTAst}{0.065}
\newcommand{\DSDivCovOne}{0.293}
\newcommand{\DSDivCovHundred}{0.632}
\newcommand{\DSDivAst}{0.308}
\newcommand{\DSChainAst}{0.325}
\newcommand{\DSGainHundred}{12.4}
\newcommand{\DSCtrlGainHundred}{-3.4}
\newcommand{\DSPSFTGainHundred}{-2.4}
\newcommand{\DSPSFTBestEpoch}{2}

\newcommand{\SweepGainMin}{7.9}
\newcommand{\SweepGainMax}{12.3}
\newcommand{\SweepContrastTotal}{12}
\newcommand{\SweepContrastShorter}{11}
\newcommand{\SweepContrastResolvedWord}{four}
\newcommand{\SweepMBFiveDelta}{+2.4}
\newcommand{\SweepMBFiveLo}{+0.8}
\newcommand{\SweepMBFiveHi}{+4.2}
\newcommand{\SweepDSThreeDelta}{+3.1}
\newcommand{\SweepDSThreeLo}{+1.1}
\newcommand{\SweepDSThreeHi}{+5.1}
\newcommand{\SweepDSFiveDelta}{+2.9}
\newcommand{\SweepDSFiveLo}{+1.2}
\newcommand{\SweepDSFiveHi}{+4.6}
\newcommand{\SweepDSSevenDelta}{+1.6}
\newcommand{\SweepDSSevenLo}{+0.3}
\newcommand{\SweepDSSevenHi}{+3.0}
\newcommand{\DSSweepOneTwo}{-4.6}
\newcommand{\DSSweepOneTen}{-7.4}

\newcommand{\HEHeldoutBase}{0.888}
\newcommand{\HEHeldoutDiv}{0.956}
\newcommand{\MBHeldoutBase}{0.751}
\newcommand{\MBHeldoutDiv}{0.881}
\newcommand{\DSHeldoutBase}{0.523}
\newcommand{\DSHeldoutDiv}{0.696}
\newcommand{\MBTransferCross}{0.871}
\newcommand{\MBTransferDelta}{-0.003}
\newcommand{\HETransferDelta}{-0.015}
\newcommand{\HENovelDiv}{18}
\newcommand{\HENovelBase}{0}
\newcommand{\MBNovelDiv}{54}
\newcommand{\MBNovelBase}{3}
\newcommand{\DSNovelDiv}{172}
\newcommand{\DSNovelBase}{34}
\newcommand{\NovelDivTotal}{244}
\newcommand{\NovelBaseUnsolved}{600}
\newcommand{\NovelRatio}{6.6}

\newcommand{\TempPeakCov}{0.910}
\newcommand{\TempPeakAst}{0.175}
\newcommand{\TempPeakValue}{2.6}
\newcommand{\TempPeakGap}{5.0}

\newcommand{\ScopeFamiliesWord}{nine}
\newcommand{\ScopeOlsSlope}{-6.14}
\newcommand{\ScopeOlsSlopeLo}{-10.18}
\newcommand{\ScopeOlsSlopeHi}{-2.10}
\newcommand{\ScopeOlsRSquared}{0.65}
\newcommand{\ScopeOlsRSquaredPct}{65}
\newcommand{\ScopeOlsP}{0.0088}
\newcommand{\ScopeConcentratedWord}{six}
\newcommand{\ScopeGainMin}{2.8}
\newcommand{\ScopeGainMax}{21.5}

\newcommand{\HEValLossOne}{0.409}
\newcommand{\HEValLossTwo}{0.389}
\newcommand{\HEValLossThree}{0.444}
\newcommand{\HEValLossFive}{0.501}
\newcommand{\HEBestEpoch}{2}
\newcommand{\HEEpochPassOne}{0.957}
\newcommand{\HEEpochPassTwo}{0.963}
\newcommand{\HEEpochPassThree}{0.955}
\newcommand{\MBBestEpoch}{1}
\newcommand{\MBEpochPassOne}{0.884}
\newcommand{\MBEpochPassTwo}{0.884}
\newcommand{\MBEpochPassThree}{0.872}
\newcommand{\MBEpochTwoDelta}{+0.012}
\newcommand{\MBEpochTwoLo}{+0.001}
\newcommand{\MBEpochTwoHi}{+0.024}
\newcommand{\DSBestEpoch}{2}

\newcommand{\PositionFirst}{0.787}
\newcommand{\PositionLast}{0.616}
\newcommand{\PositionChainLength}{20}

\newcommand{\HECorrectCandidates}{410}
\newcommand{\HECorrectRetained}{276}
\newcommand{\HECorrectUnfilteredOne}{75.7}
\newcommand{\HECorrectFilteredOne}{76.7}
\newcommand{\HECorrectDeltaOne}{+1.1}
\newcommand{\HECorrectLoOne}{+0.2}
\newcommand{\HECorrectHiOne}{+2.0}
\newcommand{\HECorrectUnfilteredHundred}{95.6}
\newcommand{\HECorrectFilteredHundred}{95.7}
\newcommand{\HECorrectDeltaHundred}{+0.1}
\newcommand{\HECorrectLoHundred}{-1.7}
\newcommand{\HECorrectHiHundred}{+1.9}
\newcommand{\MBCorrectCandidates}{945}
\newcommand{\MBCorrectRetained}{586}
\newcommand{\MBCorrectUnfilteredOne}{65.1}
\newcommand{\MBCorrectFilteredOne}{67.1}
\newcommand{\MBCorrectDeltaOne}{+2.0}
\newcommand{\MBCorrectLoOne}{+0.1}
\newcommand{\MBCorrectHiOne}{+4.1}
\newcommand{\MBCorrectUnfilteredHundred}{88.1}
\newcommand{\MBCorrectFilteredHundred}{87.4}
\newcommand{\MBCorrectDeltaHundred}{-0.6}
\newcommand{\MBCorrectLoHundred}{-2.4}
\newcommand{\MBCorrectHiHundred}{+1.0}
\newcommand{\DSCorrectCandidates}{2500}
\newcommand{\DSCorrectRetained}{689}
\newcommand{\DSCorrectUnfilteredOne}{32.0}
\newcommand{\DSCorrectFilteredOne}{33.6}
\newcommand{\DSCorrectDeltaOne}{+1.6}
\newcommand{\DSCorrectLoOne}{+0.1}
\newcommand{\DSCorrectHiOne}{+3.2}
\newcommand{\DSCorrectUnfilteredHundred}{69.6}
\newcommand{\DSCorrectFilteredHundred}{72.8}
\newcommand{\DSCorrectDeltaHundred}{+3.2}
\newcommand{\DSCorrectLoHundred}{+0.8}
\newcommand{\DSCorrectHiHundred}{+5.8}

\title{Don't Repeat Yourself: Self-Supervised Fine-Tuning for Coverage}

\author{Eric Fithian \\
  University of Chicago \\
  \texttt{efithian@uchicago.edu} \\
  \And 
  Kirill Skobelev \\
  Northwestern University \\
  \texttt{kirill@u.northwestern.edu} \\
  \And
  X.Y. Han \\
  University of Chicago \\
  \texttt{xyhan@uchicago.edu} \\
}

\begin{document}

\maketitle

\begin{abstract}
In verifiable domains, such as math and coding, where multiple solutions can be generated in parallel and scored by a verifier, getting just one correct solution among many attempts can be more important than the individual pass rate of each attempt, especially for difficult tasks where a correct solution might be a needle-in-the-haystack. Post-training methods such as supervised fine-tuning (SFT) and reinforcement learning based preference fine-tuning in Large Language Models (LLMs) have been shown to concentrate their outputs around a few modes. A common method to increase LLM output diversity is to increase a ``temperature'' parameter, but the effectiveness of this has been shown to be limited \citep{banayeeanzade2026calibration,holtzman2020curious}. To address this issue we introduce \method{} (\methodshort{}): a post-training method that increases output diversity and improves coverage --- coverage being how likely it is for there to be at least one correct solution among many generated attempts. \methodshort{} has two simple stages: first, for each problem, sequentially generate $K$ solutions, where, for each subsequent generation, the prompt contains the problem plus all of the prior problem attempts and asks for a different solution. Second, fine-tune the LLM on each solution attempt as if it was generated independently by stripping away the prior attempts from the context. This process involves no reward, verifier, or correctness filter. The result is a model that is trained to produce many diverse solutions. We evaluate \methodshort{} on three coding benchmarks and show that it raises \passkplus{100} by $\HEGainHundred$, $\MBGainHundred$, and $\DSGainHundred$ points across HumanEval+, MBPP+, and DS-1000 at a small cost to \passkplus{1}. We also measure the structural diversity using abstract syntax tree (AST) edit distance of passing solutions and find that this rises significantly across all three benchmarks. Additionally, \methodshort{} solves $\NovelDivTotal$ of $\NovelBaseUnsolved$ problems the base model was not capable of solving in the same $200$ attempts. Finally, across \ScopeFamiliesWord{} open-weight models, lower structural diversity of the base model significantly predicts higher \methodshort{} performance gains, indicating that our method is especially effective on more mode-collapsed LLMs.

\end{abstract}

\section{Introduction}
\label{sec:introduction}

In verifiable domains, such as math and coding, where multiple solutions can be generated in parallel and scored by a verifier, getting just one correct solution among many attempts can be more important than the individual pass rate of each attempt, especially for difficult tasks where a correct solution might be a needle-in-the-haystack \citep{li2022alphacode,snell2024scaling}. In this setting, a model's coverage, which is the probability that at least one of its $k$ samples is correct (\passk{k}), determines its performance. Importantly, to improve coverage, a model must generate attempts that differ meaningfully from other attempts in the context of the task objective. In other words, \passk{k} only rises if later samples solve what earlier ones couldn't \citep{brown2024monkeys}. A model that produces only a few distinct solution modes for a given problem does not benefit from large amounts of repeated sampling, resulting in \passk{k} plateauing at low $k$. Many post-training algorithms result in this exact situation, often referred to as mode-collapse. \citet{kirk2024rlhf} show that RLHF measurably sharpens and collapses the output distribution, and \citet{zhang2026verbalized} attribute this collapse to a typicality bias in preference data. \citet{yue2025limitrlvr} further show that reinforcement learning with verifiable rewards largely moves \passk{k} performance into \passk{1} without moving high \passk{k}. Thus, most post-training methods optimize for \passk{1} at the cost of \passk{k}. 

A common method for increasing output variance is to increase a ``temperature'' parameter. However, this has been shown to lack necessary control over the output distribution. \citet{banayeeanzade2026calibration} demonstrate the limitation of temperature and attribute this limitation to the characteristics of the token-level output distribution. They find that the probability mass of tokens at a given token generation step concentrates on a few valid tokens above a heavy tail that mixes valid tokens with invalid ones. This means that a temperature parameter high enough to reach the rest of the valid set also includes the invalid tokens, and a temperature low enough to stay within valid tokens limits diversity. No ranking or probability-based token decoding method can avoid this token validity and output diversity tradeoff. Decoding methods can only re-weight the distribution they are given \citep{vijayakumar2018dbs,holtzman2020curious}. Therefore, to increase the output variance without significantly degrading outputs you must change the output distribution itself instead of simply reweighting it.


To address this, we introduce \method{} (\methodshort{}, Figure~\ref{fig:method}): a post-training method that changes the output distribution by fine-tuning on a diverse set of self-generated rollouts which improves coverage and output diversity. \methodshort{} has two simple stages: first, for each problem, sequentially generate $K$ solutions, where, for each subsequent generation, the prompt contains the problem plus all of the prior problem attempts and asks for a different solution. Second, fine-tune the model on each solution attempt as if it were generated independently by removing the prior attempts from the context. By including the prior solutions in the context, the model produces a naturally diverse set of solutions. This diversity is trained into the model weights such that, at inference time, parallel independent sampling produces samples with more diversity. On three coding benchmarks \methodshort{} raises \passkplus{100} by $\HEGainHundred$, $\MBGainHundred$, and $\DSGainHundred$ points over the base model. A matched i.i.d.\ control that matches the training protocol of \methodshort{} but generates the training solutions independently moves \passkplus{100} by only $\HECtrlGainHundred$, $\MBCtrlGainHundred$, and $\DSCtrlGainHundred$ points. Additionally, P-SFT \citep{fan2026psft}, the closest published method, applied to the same self-generated solutions, moves \passkplus{100} by only $\HEPSFTGainHundred$, $\MBPSFTGainHundred$, and $\DSPSFTGainHundred$ points. Our method offers a way to significantly widen a mode-collapsed model's output distribution without a reward, a verifier, or any correctness proxy. To summarize, we make the following contributions:

\begin{figure}[t]
  \centering
  \includegraphics[width=\linewidth]{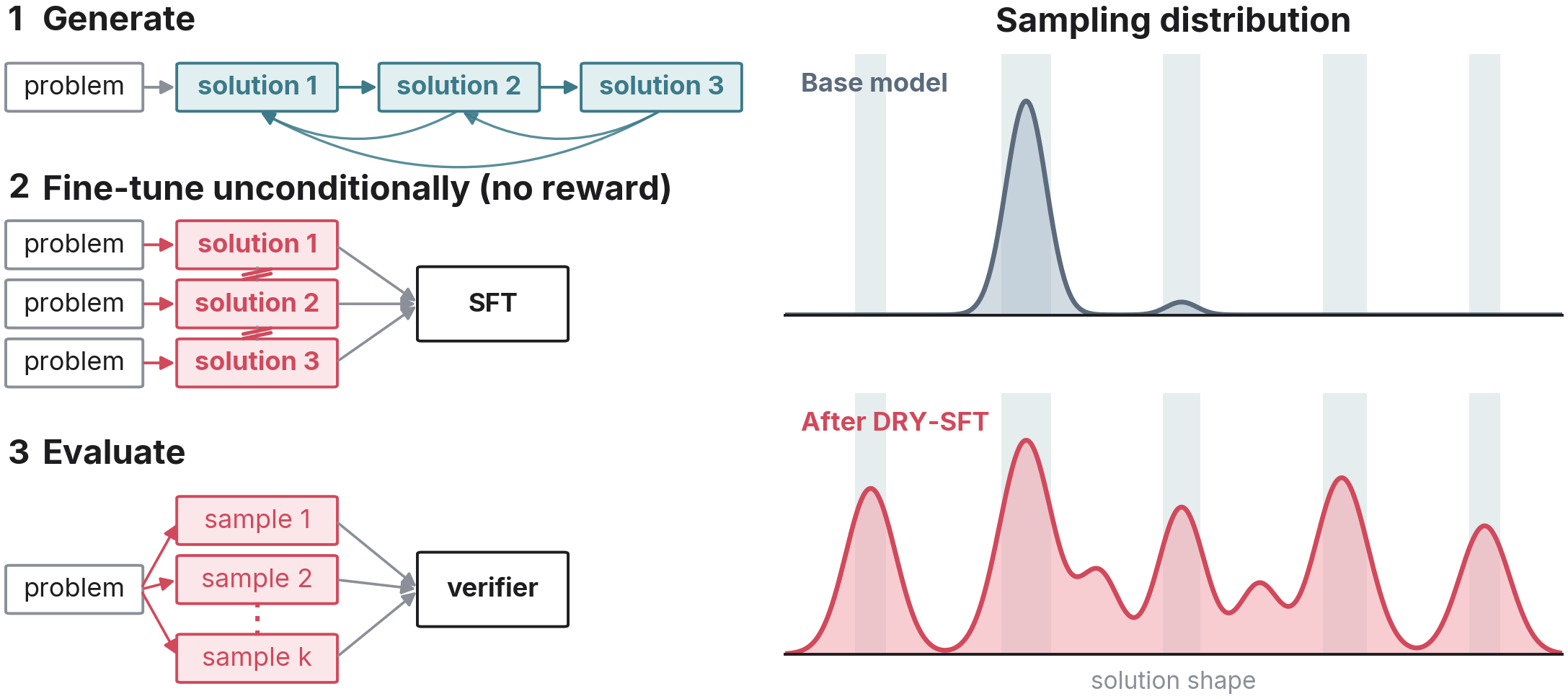}
  \caption{\methodshort{}. \textbf{Generate:} a chain of solutions per problem,
    each conditioned on those before it. \textbf{Fine-tune:} the conditioning is
    discarded and each solution is paired with the problem for
    SFT: no reward, no verifier, no filter. \textbf{Evaluate:} solutions
    are drawn independently and in parallel from the prompt.
    Right: the illustrative change in the distribution over solution types. The distributions illustrate the mechanism and are not measured. The measured
    structural change is in Figure~\ref{fig:results}.}
  \label{fig:method}
\end{figure}

\begin{enumerate}
\item \textbf{A reward-free method for increasing output diversity.} \methodshort{} is a post-training method for increasing the diversity of a model's output distribution without any reward, verifier, or filtering.

\item \textbf{A characterization of the \methodshort{} objective.} \methodshort{} learns a mixture over $K$ distinctly conditioned contexts for each problem. We contrast against a control that isolates this mechanism.

\item \textbf{Coverage and diversity gains across three code benchmarks.} \methodshort{} significantly raises \passkplus{100}, increases structural diversity, generalizes gains out-of-sample and across datasets, and solves problems unsolved by the base model across HumanEval+, MBPP+, and DS-1000. Neither the control, P-SFT \citep{fan2026psft}, nor a temperature sweep reproduces the same gains.

\item \textbf{A test to predict the effectiveness of \methodshort{}.} Across \ScopeFamiliesWord{} open-weight models, the structural diversity of the base model correlates with the effectiveness of \methodshort{} ($R^2{=}\ScopeOlsRSquared$, $p{=}\ScopeOlsP$).
\end{enumerate}
\section{Related Work}
\label{sec:related_work}

\paragraph{Rewarding coverage with a verifier.} The simple way to increase coverage is to target it directly. \citet{chen2025passktraining} make \passk{k} itself the reward by rewarding a group of $k$ rollouts when at least one is correct. \citet{cheng2025entropy} instead add a token-level entropy term to the advantage, encouraging exploration at the pivotal tokens that determine or connect logical steps. However, both of these methods require a verifier, whereas \methodshort{} does not.

\paragraph{Filtering self-generated data.} A different set of methods applies a correctness filter and then fine-tunes on the correct attempts instead of maximizing a reward. STaR \citep{zelikman2022star} and ReST$^{EM}$ \citep{singh2024restem} both generate solutions for each problem, filter for correctness, refit the base model on the correct set with SFT, and repeat. GiFT \citep{li2025gift} alternates between generating code from a description and generating a description of that generated code, Gibbs-style, producing a diverse set of descriptions and solutions for each problem which is then filtered for correctness and trained into the model. Here, the generations are conditioned on rephrasings of the problem, where ours is conditioned on previous solutions. Recursive training on synthetic data can truncate the tails of a model's output distribution \citep{shumailov2024collapse}, but its effects depend on whether synthetic samples replace earlier data or accumulate alongside the original data \citep{gerstgrasser2024collapse}. These results concern recursive data reuse, not correctness filtering alone. Our single-cycle experiments do not establish stability under repeated self-training.

\paragraph{Generation under privileged context.} Some methods generate solutions under a context that is removed during training. \citet{snell2022context} and \citet{yu2024system2} both prompt a model to generate a reasoning trace and an answer to a given problem and then fine-tune the model to predict its answer from the prompt alone. However, neither of these methods elicit diversity. SESA does elicit diversity \citep{kang2025sesa} by generating solution sketches sequentially such that each is conditioned on those before it. These sketches are then added to the prompt separately during reinforcement learning rollouts. Importantly, SESA requires a sequential sketch generation step before any rollouts can be generated. Our method instead amortizes this sequential generation step by training it into the model with SFT without rewards.

\paragraph{Changing the objective, or presupposing variation in the data.} Other methods aim to preserve diversity directly through fine-tuning. P-SFT \citep{fan2026psft} trains on each existing solution behind a randomly generated number prompt prefix, resampled every iteration so the prefix separates training contexts rather than acting as an index. The same prefixes also work at inference alone, without any training. SSFT \citep{jia2026forking} assigns each existing solution for a problem a reserved token and then fine-tunes on the pairings that the model is the least surprised about (lowest log-loss). SED-SFT \citep{chen2026sedsft} adds entropy regularization to the SFT objective. \citet{klypa2026sftdiversity} use a Tempered Focal loss to target the neglect of low-frequency patterns and the forgetting of pretrained knowledge that they identify as the two drivers of diversity loss. DivPO \citep{lanchantin2025divpo} builds pairs by selecting chosen responses that are rare but high quality against rejected ones that are common but low quality. Each of these methods either changes the loss or requires a set of multiple solutions per problem that it does not itself produce, whereas \methodshort{} requires neither.

\paragraph{Injecting variation during inference.} A fourth set of methods varies the context at generation time. \citet{wang2025samplingdiversity} study prompt perturbation for Best-of-$N$ scaling, deriving a diversity--fidelity trade-off that governs how much perturbation helps. \citet{zhang2026injectdiversity} propose generating diverse intermediate specifications to condition on. Verbalized Sampling \citep{zhang2026verbalized} asks the model to verbalize a probability distribution over responses. None of these require training, but because of that, they require applying the prompting method to every generation at inference time rather than training it into the model.

\paragraph{How \methodshort{} is different.} \methodshort{} is a post-training method that does not require changing the loss, optimizer, or objective and changes only how the training set is constructed. Neither a reward, a verifier, a correctness filter, nor a pre-existing multi-response corpus is required. Diverse solutions are elicited through a natural sequential generation method. This diversity is trained into the model once and results in a model where independent sampling produces meaningfully diverse solutions from a bare prompt.


\section{\method}
\label{sec:method}

\paragraph{Generation and fine-tuning.} \methodshort{} has two stages (Figure~\ref{fig:method}). In Stage~1, for each problem, $K$ solutions are generated sequentially such that each subsequent generation is prompted with the problem plus the previous generations (exact prompts are in Appendix~\ref{app:prompts}). In Stage~2, we remove the prior solutions and diversity prompt from the context. With a dataset $\mathcal{D}$ of $N$ problems, this produces $N \cdot K$ training pairs used to fine-tune the model with LoRA SFT. Neither stage uses a reward, verifier, or any kind of correctness filter, meaning that every generated solution is trained on. At test-time the model is sampled independently and in parallel using a standard prompt.

\paragraph{Objective.}
Let $\pi_\theta(y\mid x)$ be a model's distribution over solutions $y$ to a problem $x$. Stage~1 draws $K$ solutions sequentially,
\begin{equation}
  y_1 \sim \pi_\theta(\cdot\mid x), \qquad
  y_i \sim \pi_\theta(\cdot\mid x, y_{<i}), \quad i = 2,\dots,K,
  \label{eq:chain}
\end{equation}
where $y_{<i}$ contains all solution attempts on problem $x$ before generation $i$. Stage~2 then trains on each solution $y_i$ conditioned only on the problem $x$,
\begin{equation}
  \max_\theta \; \sum_{x \in \mathcal{D}} \sum_{i=1}^{K} \log \pi_\theta(y_i \mid x).
  \label{eq:sft}
\end{equation}
This fits a mixture over the $K$ conditionally generated solutions from Equation~\ref{eq:chain}. A matched i.i.d.\ control doesn't condition on $y_{<i}$, so every position is drawn from $\pi_\theta(\cdot\mid x)$ and Equation~\ref{eq:sft} fits the same distribution the model already has. This illustrates how \methodshort{} differs from plain self-generated data fine-tuning. The key step is that self-generated data is conditioned on previous attempts in Equation~\ref{eq:chain}. \citet{nguyen2026coverage} empirically motivate this. They show that when each problem has several reasoning modes, coverage is high. However, when each problem is fit to a single mode, coverage degrades. This applies even when the overall dataset is balanced across modes.

\section{Experiments}
\label{sec:experiments}

\paragraph{Data.} We evaluate \methodshort{} on three coding datasets: HumanEval+ and MBPP+ from EvalPlus \citep{liu2023evalplus} with the extended test suites, and DS-1000 ($1{,}000$ data-science problems) with its official execution harness \citep{lai2023ds1000}.

\paragraph{Coverage.} Our primary metric of interest is \passk{k} which measures a model's coverage. It is estimated as the average probability that at least one of $k$ independently generated samples is correct for any given problem in a dataset \citep{chen2021evaluating}. We report \passkplus{k}, where the plus indicates that a solution is only correct when it passes the extended EvalPlus test suite. On DS-1000 it indicates correctness on the official harness. We sample $n{=}200$ completions per problem at temperature $0.8$ and nucleus $p{=}0.95$ \citep{holtzman2020curious}, and report uncertainty with a $95\%$ percentile bootstrap over problems with $2{,}000$ resamples unless noted otherwise.

\paragraph{Structural diversity.} Coverage can only measure whether a model solves a problem within a given number of attempts, but not whether the passing solutions differ from each other in meaningful ways. To address this, we measure structural diversity in the context of code: the mean pairwise Zhang--Shasha tree edit distance \citep{zhang1989tree} between the abstract syntax trees (AST) of a problem's passing solutions, normalized by each problem's largest tree size to $[0,1]$. Comparing syntax trees lets us differentiate surface-level changes, such as different variable names, from real differences in what a program does or how it works. We measure it on the first $10$ samples per problem, over problems with at least two passing solutions.

\paragraph{Models and baselines.} We compare \methodshort{} against three baselines on Qwen3-4B-Instruct-2507 \citep{qwen2025qwen3}: the base model, a diversity-prompted i.i.d.\ SFT control, and P-SFT \citep{fan2026psft}. The three fine-tuned arms use $K{=}5$ unfiltered self-generated solutions per problem ($820$ for HumanEval+, $1{,}890$ for MBPP+, $5{,}000$ for DS-1000), the same SFT objective and LoRA optimizer, and the same validation-based checkpoint selection. P-SFT uses exactly the control generations but prepends a random state: \{\} prefix to every training pair. We follow P-SFT's unstarred protocol, which removes prefixes and uses ordinary sampling during inference. To choose $K$, we performed a hyperparameter sweep through $K{=}10$ and chose $K{=}5$ since the coverage flattened between $K\in[3,7]$ (Appendix~\ref{app:kablation}; Appendix~\ref{app:positions} reports how target quality varies along the chain). For SFT, we choose the checkpoint with the lowest cross-entropy on a $10\%$ problem-level holdout independently for each arm (Appendix~\ref{app:epoch}). Appendix~\ref{app:hyperparameters} lists all hyperparameters in more detail.


\subsection{Pass@k Performance}
\label{subsec:main}

\begin{figure}[t]
  \centering
  \includegraphics[width=\linewidth]{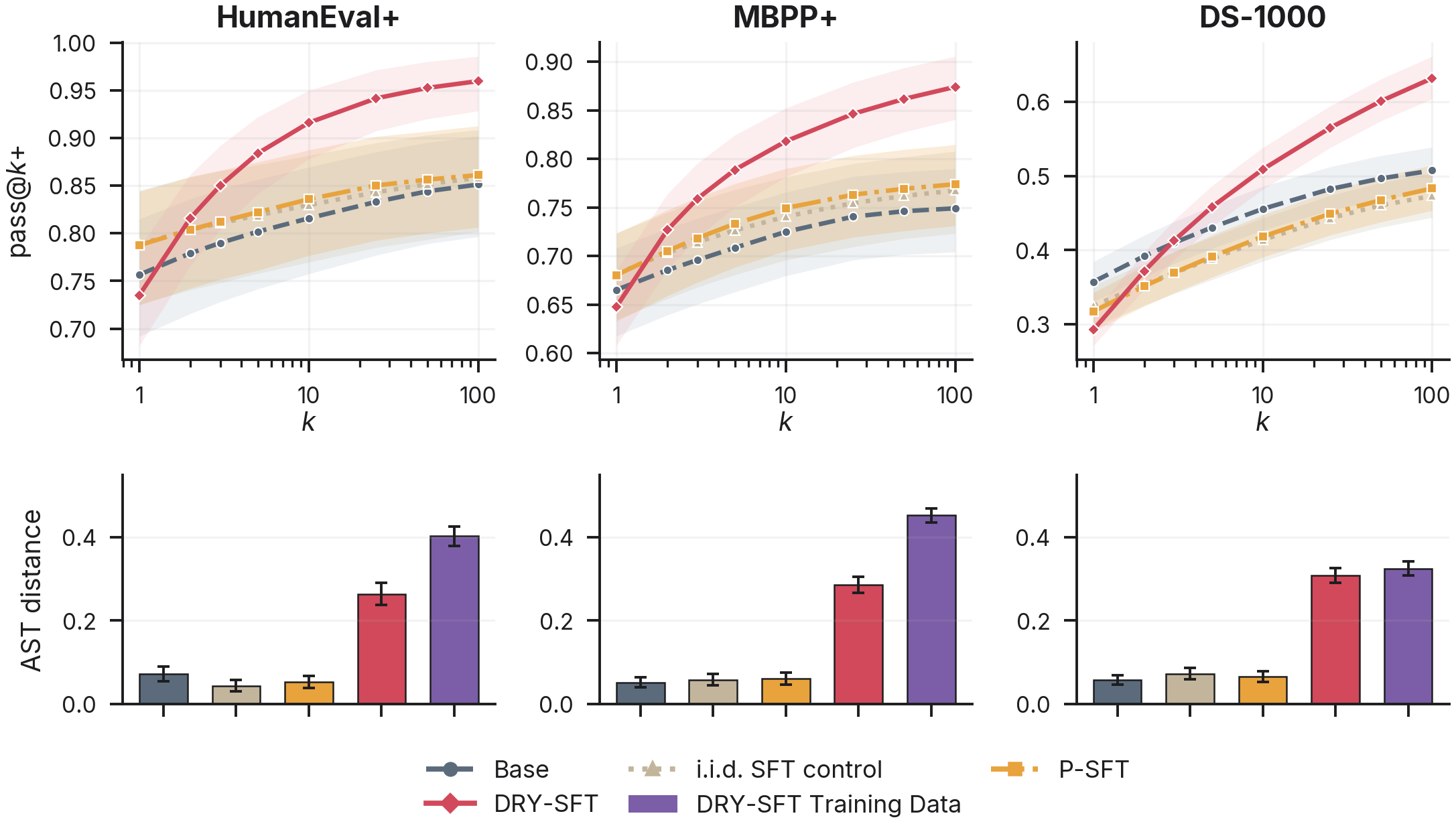}
  \caption{\methodshort{} increases coverage and structural diversity. \textbf{Top:} \passkplus{k} from $n{=}200$ independent samples per problem at $T{=}0.8$ and $p{=}0.95$. The i.i.d.\ SFT control and P-SFT use the same $5$ independently self-generated attempts; P-SFT adds resampled random-state prefixes during training but not at inference. \methodshort{} uses $5$ sequentially generated targets conditioned on prior solutions. Bands are pointwise $95\%$ problem bootstrap. \textbf{Bottom:} structural diversity, the mean pairwise tree edit distance between the abstract syntax trees (ASTs) of a problem's passing solutions, normalized to $[0,1]$ \citep{zhang1989tree}, over the first $10$ samples (Section~\ref{sec:experiments}, Structural diversity). Bars have $95\%$ bootstrapped confidence intervals over problems with at least two passing solutions. \methodshort{} Training Data is the set of Stage~1 solutions \methodshort{} was fine-tuned on, not a model arm.}
  \label{fig:results}
\end{figure}

\textbf{Coverage increases consistently across benchmarks.} 
The advantage of \methodshort{} is shown in Figure~\ref{fig:results}. \methodshort{} improves \passkplus{100} over the base model by $+\HEGainHundred$ points on HumanEval+ (from $\HEBaseCovHundred$ to $\HEDivCovHundred$), by $+\MBGainHundred$ points on MBPP+ ($\MBBaseCovHundred$ to $\MBDivCovHundred$), and by $+\DSGainHundred$ points on DS-1000 ($\DSBaseCovHundred$ to $\DSDivCovHundred$) with non-overlapping bootstrap confidence intervals. The i.i.d.\ control moves \passkplus{100} by $+\HECtrlGainHundred$, $+\MBCtrlGainHundred$, and $\DSCtrlGainHundred$ points. P-SFT moves \passkplus{100} by $+\HEPSFTGainHundred$, $+\MBPSFTGainHundred$, and $\DSPSFTGainHundred$ points.
At $k=100$, neither baseline produces the same improvement as \methodshort{} on any of the three benchmarks with non-overlapping bootstrapped 95\% confidence intervals. Randomized prefixes therefore do not substitute for conditioning on prior solutions. It is not entirely obvious whether \methodshort{} trained at $K=5$ should show improvement beyond \passkplus{5}. However, these results show that \passkplus{k} continues to improve well beyond \passkplus{5}. Additionally, Appendix~\ref{app:correctness-filtering} reports a \methodshort{} coverage ablation that compares correctness filtering versus no filtering before SFT and finds that the performance differences are relatively small.

\textbf{The \passkplus{1} trade-off.} Pass@1+ falls from $\HEBaseCovOne$ to $\HEDivCovOne$ on HumanEval+, from $\MBBaseCovOne$ to $\MBDivCovOne$ on MBPP+, and from $\DSBaseCovOne$ to $\DSDivCovOne$ on DS-1000. However, the coverage curves cross by $k{=}2$ on HumanEval+ and MBPP+ benchmarks and by $k{=}3$ on DS-1000, so \methodshort{} beats the base model after a small number of samples. 

\textbf{Solves many unsolved problems.} On HumanEval+, MBPP+, and DS-1000, \methodshort{} uniquely solves $\HENovelDiv$, $\MBNovelDiv$, and $\DSNovelDiv$ problems respectively that the base model was never able to solve within the same $200$ attempts, $\NovelDivTotal$ of the $\NovelBaseUnsolved$ problems the base model never solves. The base model uniquely solves $\HENovelBase$, $\MBNovelBase$, and $\DSNovelBase$ problems respectively, however, \methodshort{} solves roughly $\NovelRatio$ times more unique problems across the three datasets. This is a significant number of unsolved problems solved by \methodshort{} without requiring any correctness or reward signal.

\subsection{Meaningful Diversity}
\label{subsec:structure}

\textbf{Program structure increases with coverage.} Figure~\ref{fig:results} shows that the normalized AST edit distance among passing solutions rises from $\HEBaseAst$ to $\HEDivAst$ on HumanEval+, from $\MBBaseAst$ to $\MBDivAst$ on MBPP+, and from $\DSBaseAst$ to $\DSDivAst$ on DS-1000. The matched i.i.d.\ control reaches $\HECtrlAst$, $\MBCtrlAst$, and $\DSCtrlAst$, while P-SFT reaches $\HEPSFTAst$, $\MBPSFTAst$, and $\DSPSFTAst$. This supports our claim that the increase in output diversity is structural as opposed to surface level. 

\textbf{Diversity is elicited through sequential generation.} The sequentially conditioned generations that we're trained on reach $\HEChainAst$, $\MBChainAst$, and $\DSChainAst$ normalized AST edit distance, significantly above the fine-tuned model on each dataset. 
This shows that \methodshort{} takes the variation the base model could already produce, but only under a specific context, and trains it into the model's default sampling distribution. 

\subsection{Temperature}
\label{subsec:temperature}

\begin{figure}[t]
  \centering
  \includegraphics[width=\linewidth]{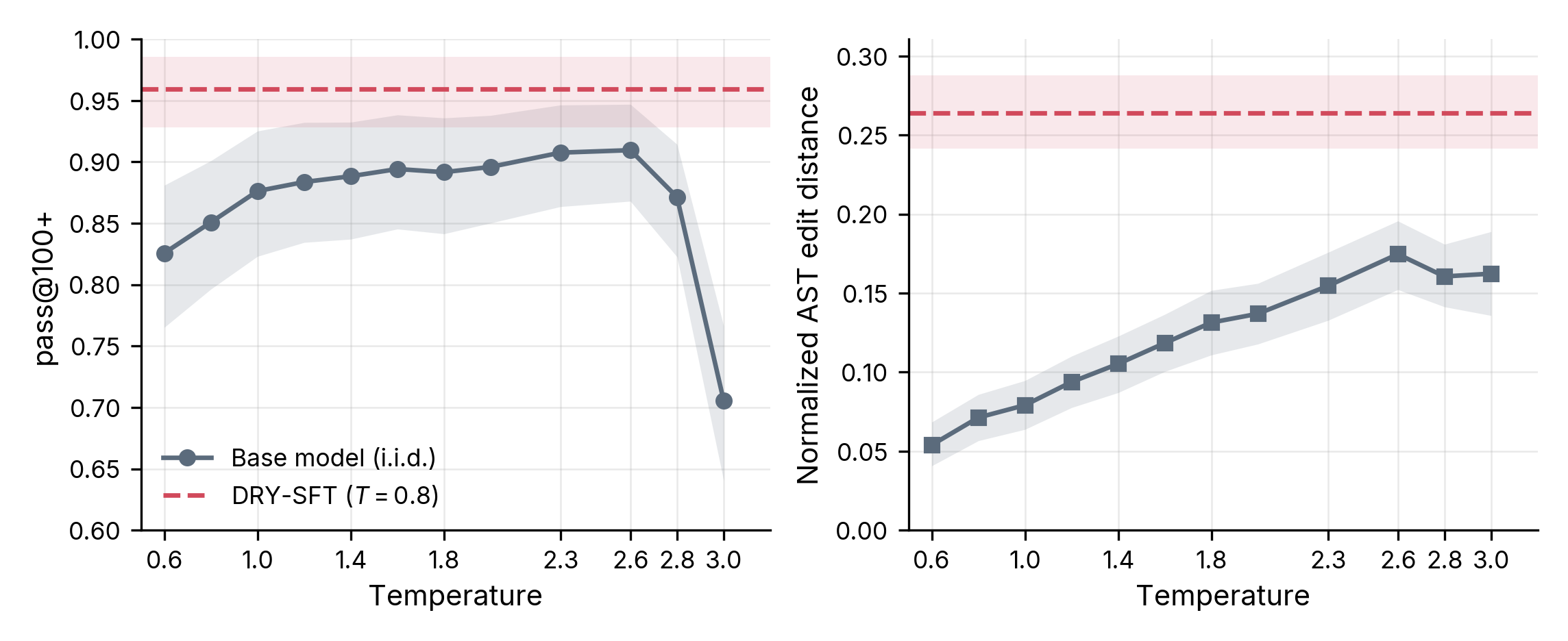}
  \caption{Temperature recovers coverage but not structure. Base-model
    \passkplus{100} (left) and structural diversity (normalized AST edit
    distance among passing solutions, defined in Section~\ref{sec:experiments})
    (right) against temperature on HumanEval+, over a grid of $12$
    temperatures from $0.6$ to $3.0$. Coverage climbs to a
    peak at $T{=}\TempPeakValue$ and falls away beyond it, while structural
    diversity reaches \methodshort{}'s level at no temperature. Shaded regions
    are pointwise $95\%$ problem-bootstrap confidence bands; dashed lines and
    bands mark \methodshort{} at $T{=}0.8$.}
  \label{fig:temp}
\end{figure}

\textbf{Increasing temperature recovers some coverage but not structural diversity.} In Figure~\ref{fig:temp}, we evaluate whether raising the base model's sampling temperature reproduces the coverage and diversity gains of \methodshort{}. The best temperature on the grid reaches \passkplus{100} $=\TempPeakCov$ at $T{=}\TempPeakValue$ on HumanEval+. \methodshort{} outperforms this by $\TempPeakGap$ points. However, the intervals overlap, so we read this as a point-estimate difference rather than a statistically significant difference. When comparing the structural diversity at that peak temperature, the normalized AST edit distance is $\TempPeakAst$, against $\HEDivAst$ for \methodshort{}. \methodshort{} clearly has a more structurally diverse output distribution. Also, notice that there is a steep dropoff near $T{=}3.0$ for the base model. This highlights the issue that \citet{banayeeanzade2026calibration} identify when temperature is too high: the sampling distribution starts to put too much mass on invalid tokens, and the model no longer reliably produces valid generations. \methodshort{} avoids this failure mode and significantly increases both output diversity and coverage.

\subsection{Held-out and Cross-dataset Performance}
\label{subsec:transfer}

\begin{figure}[t]
  \centering
  \includegraphics[width=\linewidth]{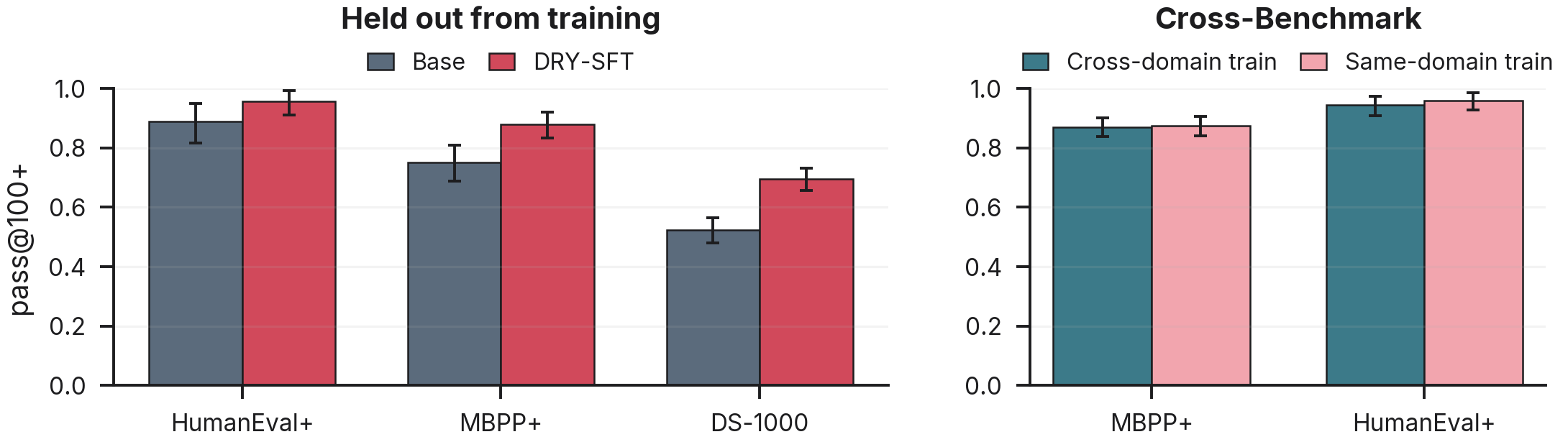}
  \caption{The gain is not tied to the problems the targets came from.
    \textbf{Left:} \passkplus{100} on the held-out half of a 50/50 split, with
    \methodshort{} trained only on the other half. \textbf{Right:}
    \methodshort{} trained on the opposite EvalPlus benchmark against
    same-domain training, evaluated on the benchmark named. Error bars are
    $95\%$ problem bootstrap.}
  \label{fig:generalization}
\end{figure}

\textbf{Performance transfers on a held-out set.} The experiments in Sections~\ref{subsec:main} and~\ref{subsec:structure} (Figure~\ref{fig:results}) generate solutions to and evaluate on the same problems. A dataset split is not an important concern in this setting since there is no correctness signal used for training. However, we address any leakage concerns in this experiment by training on one half and evaluating on the other for each of the three datasets. Figure~\ref{fig:generalization} shows \passkplus{100} on the held-out set increases from $\HEHeldoutBase$ to $\HEHeldoutDiv$ on HumanEval+, from $\MBHeldoutBase$ to $\MBHeldoutDiv$ on MBPP+, and from $\DSHeldoutBase$ to $\DSHeldoutDiv$ on DS-1000, ruling out leakage or problem memorization.

\textbf{Performance transfers across datasets.} We further investigate how well \methodshort{} generalizes coverage across datasets. Figure~\ref{fig:generalization} shows \methodshort{} trained on HumanEval+ and evaluated on MBPP+ reaches \passkplus{100} $=\MBTransferCross$, only a $\MBTransferDelta$ difference compared to the \methodshort{} model trained on MBPP+. Similarly, the reverse training and evaluation setup reveals only a $\HETransferDelta$ difference compared to the in-domain HumanEval+ trained model. This demonstrates that the increased coverage and diversity of \methodshort{} can generalize across datasets. 

\subsection{When \methodshort{} works}
\label{subsec:scope}

\begin{figure}[t]
  \centering
  \includegraphics[width=\linewidth]{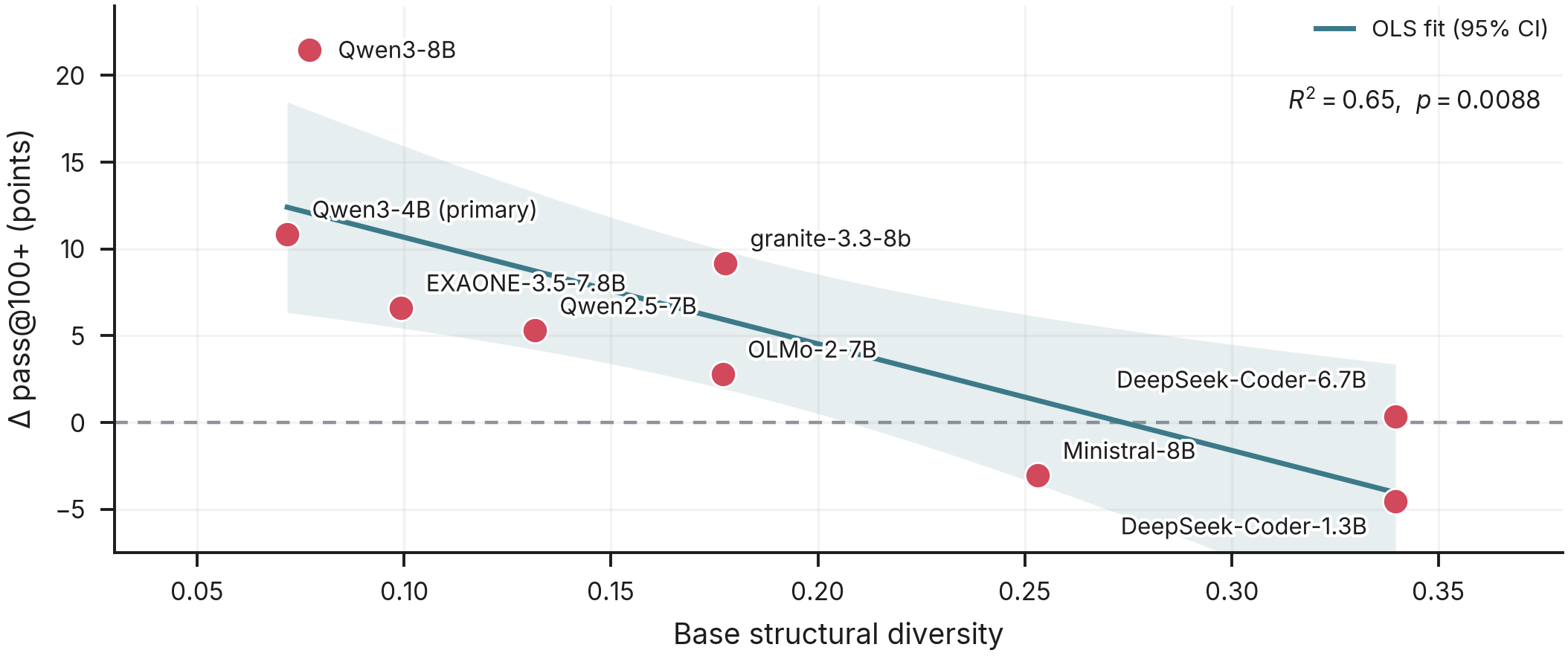}
  \caption{The gain tracks how concentrated the base model already is. Each point is one fine-tuned model: its base structural diversity --- the normalized AST edit distance among its passing solutions --- against the change in \passkplus{100} after \methodshort{}. The line is an ordinary least-squares fit over \ScopeFamiliesWord{} runs with a $95\%$ confidence band; $R^2$ and $p$ are printed in the panel. The fitted slope is $\ScopeOlsSlope$ points per $0.1$ of base diversity, with $95\%$ CI $[\ScopeOlsSlopeLo,\ScopeOlsSlopeHi]$. Models below the dashed line lose coverage. Every model is measured on HumanEval+.}
  \label{fig:scope}
\end{figure}

\textbf{Base model diversity predicts method gains.} We applied \methodshort{} to \ScopeFamiliesWord{} different open-weight models \citep{qwen2025qwen3,lgai2024exaone35,qwen2024qwen25,granite2025granite33,olmo2024olmo2,mistral2024ministraux,guo2024deepseekcoder} and plotted each model's change in \passkplus{100} against the normalized AST edit distance of its base model (Figure~\ref{fig:scope}). The base structural diversity explains $\ScopeOlsRSquaredPct$\% of the variation in $\Delta$ pass@100+ points across the models ($R^2{=}\ScopeOlsRSquared$, $p{=}\ScopeOlsP$). The \ScopeConcentratedWord{} models with the lowest base structural diversity all gain between $\ScopeGainMin$ and $\ScopeGainMax$ $\Delta$ pass@100+ points. This indicates that the more mode collapsed a model is, the more it will benefit from \methodshort{}. This also motivates a practical test to run before applying \methodshort{} to estimate the coverage gains.


\section{Conclusion}
\label{sec:conclusion}

We introduced a new post-training method that improves coverage and introduces diversity into mode-collapsed LLMs. We have shown that this method significantly improves pass@100+ on three coding benchmarks while increasing meaningful, structural diversity of the output distribution. We have demonstrated that increasing temperature does not fully recover these gains and that continuing to raise temperature eventually degrades pass@100+. We have also established that \methodshort{} transfers performance and diversity across benchmarks. Lastly, we identified a significant relationship between base model diversity and the effectiveness of \methodshort{}, providing a practical LLM diversity test to estimate the coverage gains of \methodshort{}.


\paragraph{Limitations.} \methodshort{} causes a decrease in \passkplus{1} across all datasets. However, the primary motivation of our method is to raise pass@k. Second, \methodshort{} did not improve coverage for models whose baseline output distribution is already relatively diverse. However, we position \methodshort{} as a remedy for mode-collapsed models which are by definition under-diverse. Third, Stage~1 generation is sequential: the $i$-th solution cannot begin until the first $i-1$ are complete. This also means that the context grows with each addition, so eliciting $K$ training targets costs more than drawing $K$ independent samples. However, this happens only once during data generation. There is no cost at inference. We also did not evaluate on any domains outside of coding. Future work should target this limitation. Lastly, our only external baseline is P-SFT. This is because we do not compare against reward-based coverage methods which require a verifier that \methodshort{} does not.

\paragraph{Future directions.} A natural question is whether the elicited variation is present in the weights of the base model or whether the base model computes the variation over the course of sequential generation. We leave this direction open for future work. Answering it could explain both why \methodshort{} works and how its mechanism could be optimized for further improvements.

\methodshort{} significantly raises \passk{k}, and it is well documented that reinforcement learning moves \passk{k} to \passk{1} \citep{yue2025limitrlvr}. Hence, reinforcement learning could serve as a natural complement to \methodshort{}, either after \methodshort{} or alternating between the two.

A third direction could explore controlling the type of output diversity. Temperature is the traditional scalar used for controlling variation, but by tuning the sequential rollout prompt of \methodshort{} to differentiate later solutions along an axis of interest, such as lexically versus semantically for example, the model might learn to spread its output distribution along that axis.


\subsection*{AI use statement}
In this work, we used generative AI tools to support data analysis, implement methods, and polish writing. We take responsibility for the final content of this work, including text, claims, or artifacts produced with the aid of generative AI.

\bibliographystyle{\venuebst}
\bibliography{references}

@article{chen2021evaluating,
  title   = {Evaluating Large Language Models Trained on Code},
  author  = {Chen, Mark and Tworek, Jerry and Jun, Heewoo and Yuan, Qiming
             and Pinto, Henrique Ponde de Oliveira and Kaplan, Jared
             and Edwards, Harri and Burda, Yuri and Joseph, Nicholas
             and Brockman, Greg and Ray, Alex and Puri, Raul
             and Krueger, Gretchen and Petrov, Michael and Khlaaf, Heidy
             and Sastry, Girish and Mishkin, Pamela and Chan, Brooke
             and Gray, Scott and Ryder, Nick and Pavlov, Mikhail
             and Power, Alethea and others},
  journal = {arXiv preprint arXiv:2107.03374},
  year    = {2021}
}

@article{brown2024monkeys,
  title   = {Large Language Monkeys: Scaling Inference Compute
             with Repeated Sampling},
  author  = {Brown, Bradley and Juravsky, Jordan and Ehrlich, Ryan
             and Clark, Ronald and Le, Quoc V. and R\'e, Christopher
             and Mirhoseini, Azalia},
  journal = {arXiv preprint arXiv:2407.21787},
  year    = {2024}
}

@article{snell2024scaling,
  title   = {Scaling {LLM} Test-Time Compute Optimally can be More Effective
             than Scaling Model Parameters},
  author  = {Snell, Charlie and Lee, Jaehoon and Xu, Kelvin and Kumar, Aviral},
  journal = {arXiv preprint arXiv:2408.03314},
  year    = {2024}
}

@inproceedings{liu2023evalplus,
  title     = {Is Your Code Generated by {ChatGPT} Really Correct?
               Rigorous Evaluation of Large Language Models for Code Generation},
  author    = {Liu, Jiawei and Xia, Chunqiu Steven and Wang, Yuyao and Zhang, Lingming},
  booktitle = {Advances in Neural Information Processing Systems},
  year      = {2023}
}

@inproceedings{holtzman2020curious,
  title     = {The Curious Case of Neural Text Degeneration},
  author    = {Holtzman, Ari and Buys, Jan and Du, Li and Forbes, Maxwell
               and Choi, Yejin},
  booktitle = {International Conference on Learning Representations},
  year      = {2020}
}

@article{kang2025sesa,
  title   = {The Road Less Traveled: Enhancing Exploration in {LLMs}
             via Sequential Sampling},
  author  = {Kang, Shijia and Zhang, Muhan},
  journal = {arXiv preprint arXiv:2510.15502},
  year    = {2025}
}

@article{chen2026sedsft,
  title   = {{SED-SFT}: Selectively Encouraging Diversity in
             Supervised Fine-Tuning},
  author  = {Chen, Yijie and Liu, Yijin and Meng, Fandong},
  journal = {arXiv preprint arXiv:2602.07464},
  year    = {2026}
}

@inproceedings{zelikman2022star,
  title     = {{STaR}: Bootstrapping Reasoning with Reasoning},
  author    = {Zelikman, Eric and Wu, Yuhuai and Mu, Jesse and Goodman, Noah D.},
  booktitle = {Advances in Neural Information Processing Systems},
  year      = {2022}
}

@inproceedings{yue2025limitrlvr,
  title     = {Does Reinforcement Learning Really Incentivize Reasoning Capacity
               in {LLMs} Beyond the Base Model?},
  author    = {Yue, Yang and Chen, Zhiqi and Lu, Rui and Zhao, Andrew
               and Wang, Zhaokai and Yue, Yang and Song, Shiji and Huang, Gao},
  booktitle = {Advances in Neural Information Processing Systems},
  year      = {2025}
}

@article{cheng2025entropy,
  title   = {Reasoning with Exploration: An Entropy Perspective},
  author  = {Cheng, Daixuan and Huang, Shaohan and Zhu, Xuekai and Dai, Bo
             and Zhao, Wayne Xin and Zhang, Zhenliang and Wei, Furu},
  journal = {arXiv preprint arXiv:2506.14758},
  year    = {2025}
}

@article{li2022alphacode,
  title     = {Competition-Level Code Generation with {AlphaCode}},
  author    = {Li, Yujia and Choi, David and Chung, Junyoung and Kushman, Nate
               and Schrittwieser, Julian and Leblond, R\'emi and Eccles, Tom
               and Keeling, James and Gimeno, Felix and Dal Lago, Agustin
               and Hubert, Thomas and Choy, Peter and de Masson d'Autume, Cyprien
               and Babuschkin, Igor and Chen, Xinyun and Huang, Po-Sen
               and Welbl, Johannes and Gowal, Sven and Cherepanov, Alexey
               and Molloy, James and Mankowitz, Daniel J. and others},
  journal   = {Science},
  volume    = {378},
  number    = {6624},
  pages     = {1092--1097},
  year      = {2022}
}

@article{wang2025samplingdiversity,
  title   = {On the Effect of Sampling Diversity in Scaling {LLM} Inference},
  author  = {Wang, Tianchun and Liu, Zichuan and Chen, Yuanzhou
             and Light, Jonathan and Liu, Weiyang and Chen, Haifeng
             and Zhang, Xiang and Cheng, Wei},
  journal = {arXiv preprint arXiv:2502.11027},
  year    = {2025}
}

@inproceedings{kirk2024rlhf,
  title     = {Understanding the Effects of {RLHF} on {LLM} Generalisation
               and Diversity},
  author    = {Kirk, Robert and Mediratta, Ishita and Nalmpantis, Christoforos
               and Luketina, Jelena and Hambro, Eric and Grefenstette, Edward
               and Raileanu, Roberta},
  booktitle = {International Conference on Learning Representations},
  note      = {Preprint: arXiv:2310.06452},
  year      = {2024}
}

@inproceedings{zhang2026verbalized,
  title     = {Verbalized Sampling: How to Mitigate Mode Collapse and
               Unlock {LLM} Diversity},
  author    = {Zhang, Jiayi and Yu, Simon and Chong, Derek and Sicilia, Anthony
               and Tomz, Michael R. and Manning, Christopher D. and Shi, Weiyan},
  booktitle = {International Conference on Machine Learning},
  year      = {2026}
}

@article{banayeeanzade2026calibration,
  title   = {Sampling More, Getting Less: Calibration is the Diversity
             Bottleneck in {LLMs}},
  author  = {Banayeeanzade, Amin and Yang, Qingchuan and Tarsadiya, Dhruv
             and Bahrani, Fatemeh and Blas, Leonardo and Samuel, Alfy
             and Jia, Robin and Razaviyayn, Meisam
             and Karimireddy, Sai Praneeth},
  journal = {arXiv preprint arXiv:2605.11128},
  year    = {2026}
}

@inproceedings{vijayakumar2018dbs,
  title     = {Diverse Beam Search for Improved Description of
               Complex Scenes},
  author    = {Vijayakumar, Ashwin K and Cogswell, Michael
               and Selvaraju, Ramprasaath R. and Sun, Qing and Lee, Stefan
               and Crandall, David and Batra, Dhruv},
  booktitle = {AAAI Conference on Artificial Intelligence},
  note      = {Preprint: arXiv:1610.02424},
  year      = {2018}
}

@article{zhang2026injectdiversity,
  title   = {Where You Inject Diversity Matters: A Unified Framework for
             Diverse Generation},
  author  = {Zhang, Cheng and Xin, Rui and Zhong, Chudi},
  journal = {arXiv preprint arXiv:2606.10302},
  year    = {2026}
}

@article{nguyen2026coverage,
  title   = {Why Do Reasoning Models Lose Coverage? The Role of Data and
             Forks in the Road},
  author  = {Nguyen, Ngoc-Hieu and Shojaee, Parshin and Nguyen, Phuc Minh
             and Zhang, Nan and Reddy, Chandan K. and Doan, Khoa D.
             and Zhang, Rui},
  journal = {arXiv preprint arXiv:2605.17026},
  year    = {2026}
}

@inproceedings{fan2026psft,
  title     = {Learning Diverse Responses with Prefix-Conditioned
               Supervised Fine-Tuning},
  author    = {Fan, Zhiyuan and Chen, Guanqiao and Huang, Yanyi
               and Zhao, Mingkuan and Guo, Dadi and Fung, Yi R.},
  booktitle = {Proceedings of the 64th Annual Meeting of the Association for
               Computational Linguistics (Volume 1: Long Papers)},
  pages     = {247--276},
  year      = {2026}
}

@inproceedings{jia2026forking,
  title     = {Training Large Language Models to Reason in Parallel with
               Global Forking Tokens},
  author    = {Jia, Sheng and Wang, Xiao and Kasiviswanathan, Shiva Prasad},
  booktitle = {International Conference on Learning Representations},
  note      = {Preprint: arXiv:2510.05132},
  year      = {2026}
}

@article{klypa2026sftdiversity,
  title   = {Diversity in Large Language Models under Supervised Fine-Tuning},
  author  = {Klypa, Roman and Cherednichenko, Oleksandr},
  journal = {arXiv preprint arXiv:2605.00195},
  year    = {2026}
}

@article{snell2022context,
  title   = {Learning by Distilling Context},
  author  = {Snell, Charlie and Klein, Dan and Zhong, Ruiqi},
  journal = {arXiv preprint arXiv:2209.15189},
  year    = {2022}
}

@article{yu2024system2,
  title   = {Distilling System 2 into System 1},
  author  = {Yu, Ping and Xu, Jing and Weston, Jason and Kulikov, Ilia},
  journal = {arXiv preprint arXiv:2407.06023},
  year    = {2024}
}

@article{singh2024restem,
  title   = {Beyond Human Data: Scaling Self-Training for Problem-Solving
             with Language Models},
  author  = {Singh, Avi and Co-Reyes, John D. and Agarwal, Rishabh
             and Anand, Ankesh and Patil, Piyush and Garcia, Xavier
             and Liu, Peter J. and Harrison, James and Lee, Jaehoon
             and Xu, Kelvin and Parisi, Aaron and Kumar, Abhishek
             and Alemi, Alex and others},
  journal = {Transactions on Machine Learning Research},
  year    = {2024}
}

@article{shumailov2024collapse,
  title   = {{AI} Models Collapse When Trained on Recursively Generated Data},
  author  = {Shumailov, Ilia and Shumaylov, Zakhar and Zhao, Yiren
             and Papernot, Nicolas and Anderson, Ross and Gal, Yarin},
  journal = {Nature},
  volume  = {631},
  number  = {8022},
  pages   = {755--759},
  year    = {2024}
}

@article{gerstgrasser2024collapse,
  title   = {Is Model Collapse Inevitable? Breaking the Curse of Recursion by
             Accumulating Real and Synthetic Data},
  author  = {Gerstgrasser, Matthias and Schaeffer, Rylan and Dey, Apratim
             and Rafailov, Rafael and Sleight, Henry and Hughes, John
             and Korbak, Tomasz and Agrawal, Rajashree and Pai, Dhruv
             and Gromov, Andrey and Roberts, Daniel A. and Yang, Diyi
             and Donoho, David L. and Koyejo, Sanmi},
  journal = {arXiv preprint arXiv:2404.01413},
  year    = {2024}
}

@inproceedings{li2025gift,
  title     = {{GiFT}: Gibbs Fine-Tuning for Code Generation},
  author    = {Li, Haochen and Feng, Wanjin and Zhou, Xin and Shen, Zhiqi},
  booktitle = {Proceedings of the 63rd Annual Meeting of the Association for
               Computational Linguistics},
  year      = {2025}
}

@article{lanchantin2025divpo,
  title   = {Diverse Preference Optimization},
  author  = {Lanchantin, Jack and Chen, Angelica and Dhuliawala, Shehzaad
             and Yu, Ping and Weston, Jason and Sukhbaatar, Sainbayar
             and Kulikov, Ilia},
  journal = {arXiv preprint arXiv:2501.18101},
  year    = {2025}
}

@article{chen2025passktraining,
  title   = {Pass@k Training for Adaptively Balancing Exploration and
             Exploitation of Large Reasoning Models},
  author  = {Chen, Zhipeng and Qin, Xiaobo and Wu, Youbin and Ling, Yue
             and Ye, Qinghao and Zhao, Wayne Xin and Shi, Guang},
  journal = {arXiv preprint arXiv:2508.10751},
  year    = {2025}
}

@article{qwen2025qwen3,
  title   = {{Qwen3} Technical Report},
  author  = {{Qwen Team}},
  journal = {arXiv preprint arXiv:2505.09388},
  year    = {2025}
}

@inproceedings{lai2023ds1000,
  title     = {{DS}-1000: A Natural and Reliable Benchmark for Data Science
               Code Generation},
  author    = {Lai, Yuhang and Li, Chengxi and Wang, Yiming and Zhang, Tianyi
               and Zhong, Ruiqi and Zettlemoyer, Luke and Yih, Wen-Tau
               and Fried, Daniel and Wang, Sida and Yu, Tao},
  booktitle = {International Conference on Machine Learning},
  pages     = {18319--18345},
  year      = {2023}
}

@article{qwen2024qwen25,
  title   = {{Qwen2.5} Technical Report},
  author  = {{Qwen Team}},
  journal = {arXiv preprint arXiv:2412.15115},
  year    = {2024}
}

@article{lgai2024exaone35,
  title   = {{EXAONE} 3.5: Series of Large Language Models for
             Real-world Use Cases},
  author  = {{LG AI Research}},
  journal = {arXiv preprint arXiv:2412.04862},
  year    = {2024}
}

@article{olmo2024olmo2,
  title   = {2 {OLMo} 2 Furious},
  author  = {{Team OLMo} and Walsh, Pete and Soldaini, Luca
             and Groeneveld, Dirk and Lo, Kyle and Arora, Shane
             and Bhagia, Akshita and Gu, Yuling and Huang, Shengyi
             and Jordan, Matt and Lambert, Nathan and Schwenk, Dustin
             and Tafjord, Oyvind and Anderson, Taira and Atkinson, David
             and Brahman, Faeze and Clark, Christopher and Dasigi, Pradeep
             and Dziri, Nouha and Guerquin, Michal and Ivison, Hamish
             and Koh, Pang Wei and Liu, Jiacheng and Malik, Saumya
             and Merrill, William and Miranda, Lester James V.
             and Morrison, Jacob and Murray, Tyler and Nam, Crystal
             and Pyatkin, Valentina and Rangapur, Aman and Schmitz, Michael
             and Skjonsberg, Sam and Wadden, David and Wilhelm, Christopher
             and Wilson, Michael and Zettlemoyer, Luke and Farhadi, Ali
             and Smith, Noah A. and Hajishirzi, Hannaneh},
  journal = {arXiv preprint arXiv:2501.00656v1},
  year    = {2024}
}

@article{guo2024deepseekcoder,
  title   = {{DeepSeek-Coder}: When the Large Language Model Meets
             Programming --- The Rise of Code Intelligence},
  author  = {Guo, Daya and Zhu, Qihao and Yang, Dejian and Xie, Zhenda
             and Dong, Kai and Zhang, Wentao and Chen, Guanting
             and Bi, Xiao and Wu, Y. and Li, Y. K. and Luo, Fuli
             and Xiong, Yingfei and Liang, Wenfeng},
  journal = {arXiv preprint arXiv:2401.14196},
  year    = {2024}
}

@misc{granite2025granite33,
  title        = {{Granite-3.3-8B-Instruct}},
  author       = {{Granite Team, IBM}},
  year         = {2025},
  howpublished = {\url{https://huggingface.co/ibm-granite/granite-3.3-8b-instruct}}
}

@misc{mistral2024ministraux,
  title        = {Un Ministral, des Ministraux},
  author       = {{Mistral AI}},
  year         = {2024},
  howpublished = {\url{https://mistral.ai/news/ministraux/}}
}

@article{zhang1989tree,
  title   = {Simple Fast Algorithms for the Editing Distance between Trees
             and Related Problems},
  author  = {Zhang, Kaizhong and Shasha, Dennis},
  journal = {SIAM Journal on Computing},
  volume  = {18},
  number  = {6},
  pages   = {1245--1262},
  year    = {1989}
}

\appendix
\counterwithin{figure}{section}
\counterwithin{table}{section}
\renewcommand{\topfraction}{0.9}
\renewcommand{\bottomfraction}{0.6}
\renewcommand{\textfraction}{0.05}
\renewcommand{\floatpagefraction}{0.8}
\setcounter{topnumber}{3}
\setcounter{totalnumber}{4}
\section{Hyperparameters}
\label{app:hyperparameters}

The generation and fine-tuning settings are shared by \methodshort{}, P-SFT, and the matched i.i.d.\ SFT control. P-SFT and the control use identical independent targets and differ only in randomized-prefix conditioning during training; \methodshort{} changes how those targets are generated. Table~\ref{tab:hyperparameters} lists every setting used by the reported runs.

\begin{table}[ht]
  \centering
  \fontsize{10}{11}\selectfont
  \caption{Every hyperparameter of the reported runs. Appendix~\ref{app:epoch} covers how the epoch is selected and Appendix~\ref{app:kablation} the choice of $K$.}
  \label{tab:hyperparameters}
  \begin{tabularx}{\linewidth}{@{}ll>{\raggedright\arraybackslash}X@{}}
    \toprule
    & Setting & Value \\
    \midrule
    \multirow{4}{*}{Generation}
      & model & Qwen3-4B-Instruct-2507 \\
      & solutions per problem, $K$ & $5$ \\
      & temperature / nucleus $p$ & $0.8$ / $0.95$ \\
      & correctness filtering & none \\
    \midrule
    \multirow{15}{*}{Fine-tuning}
      & method & LoRA \\
      & LoRA rank / scaling $\alpha$ & $64$ / $128$ \\
      & LoRA dropout & $0.05$ \\
      & LoRA target modules & \texttt{q\_proj}, \texttt{k\_proj}, \texttt{v\_proj}, \texttt{o\_proj}, \texttt{gate\_proj}, \texttt{up\_proj}, \texttt{down\_proj} \\
      & optimizer & PyTorch AdamW \\
      & AdamW $\beta_1$ / $\beta_2$ / $\epsilon$ & $0.9$ / $0.999$ / $10^{-8}$ \\
      & weight decay & $0$ \\
      & learning rate & $2{\times}10^{-5}$ \\
      & learning-rate schedule & cosine decay with $10\%$ warmup \\
      & maximum gradient norm & $1.0$ \\
      & batch size & $16$ \\
      & maximum sequence length & $8{,}192$ tokens \\
      & epoch budget & $5$ \\
      & checkpoint selection & lowest cross-entropy on a $10\%$ problem-level holdout \\
      & selected epoch & $\HEBestEpoch$ (HumanEval+), $\MBBestEpoch$ (MBPP+), $\DSBestEpoch$ (DS-1000) \\
    \midrule
    \multirow{4}{*}{P-SFT}
      & prefix template & \texttt{random state: \{00000--99999\}} followed by a space and two newlines \\
      & training prefix draw & without replacement; resampled every presentation \\
      & selected epoch & $\HEPSFTBestEpoch$ (HumanEval+), $\MBPSFTBestEpoch$ (MBPP+), $\DSPSFTBestEpoch$ (DS-1000) \\
      & plotted inference & no prefix; temperature $0.8$, nucleus $p{=}0.95$ \\
    \midrule
    \multirow{4}{*}{Evaluation}
      & samples per problem, $n$ & $200$ \\
      & temperature / nucleus $p$ & $0.8$ / $0.95$ \\
      & \passkplus{k} estimator & unbiased \citep{chen2021evaluating} \\
      & uncertainty & $95\%$ percentile bootstrap over problems, $2{,}000$ resamples \\
    \midrule
    \multirow{3}{*}{Training examples}
      & HumanEval+ & $820$ \\
      & MBPP+ & $1{,}890$ \\
      & DS-1000 & $5{,}000$ \\
    \bottomrule
  \end{tabularx}
\end{table}

\section{Choosing the training epoch}
\label{app:epoch}

For each fine-tuned arm, we hold out $10\%$ of problems and train for five epochs, restoring the checkpoint with the lowest held-out cross-entropy. We split by problem rather than example because the $K$ targets of a problem share a prompt, and an example-level split would put near-duplicates on both sides of the split.

\textbf{A fixed three-epoch budget trains too long.} To confirm that validation loss is not merely tracking likelihood, we score intermediate checkpoints directly. On HumanEval+ at $K{=}5$, \passkplus{100} is $\HEEpochPassOne$, $\HEEpochPassTwo$, and $\HEEpochPassThree$ at epochs 1, 2, and 3, respectively. On MBPP+, \passkplus{100} is $\MBEpochPassOne$, $\MBEpochPassTwo$, and $\MBEpochPassThree$ at epochs 1, 2, and 3, respectively. The fixed epoch-3 checkpoint is the worst of the three on both benchmarks. On MBPP+, the advantage of epoch 2 over epoch 3 is $\MBEpochTwoDelta$, with a paired problem-bootstrap interval of $[\MBEpochTwoLo,\MBEpochTwoHi]$. We also find that the validation loss reaches its minimum before the third epoch on every benchmark and rises monotonically thereafter. For example, on HumanEval+ the validation loss is $\HEValLossOne$, $\HEValLossTwo$, $\HEValLossThree$, and $\HEValLossFive$ at epochs 1, 2, 3, and 5, respectively.

\textbf{The selection is robust to the holdout size.} Repeating the procedure with a $30\%$ holdout ($49$ validation problems on HumanEval+ instead of $16$) selects the same epoch on every arm. The epoch that minimizes the validation loss is also among the epochs that maximize \passkplus{100} of the epochs we scored. This matters because the goal is to spread across solution structures, which cross-entropy does not measure.

\textbf{What this does not settle.} First, the number of epochs is not fully separable from $K$, since the number of updates scales with the size of the training set (Appendix~\ref{app:kablation}). Second, the selected epoch is not a property of the benchmark alone: the held-out arms select later epochs than their main counterparts (3 against 2 on HumanEval+, and 2 against 1 on MBPP+), as expected when training on half as many problems.

\section{Choosing the chain depth \texorpdfstring{$K$}{K}}
\label{app:kablation}

\begin{figure}[ht]
  \centering
  \includegraphics[width=\linewidth]{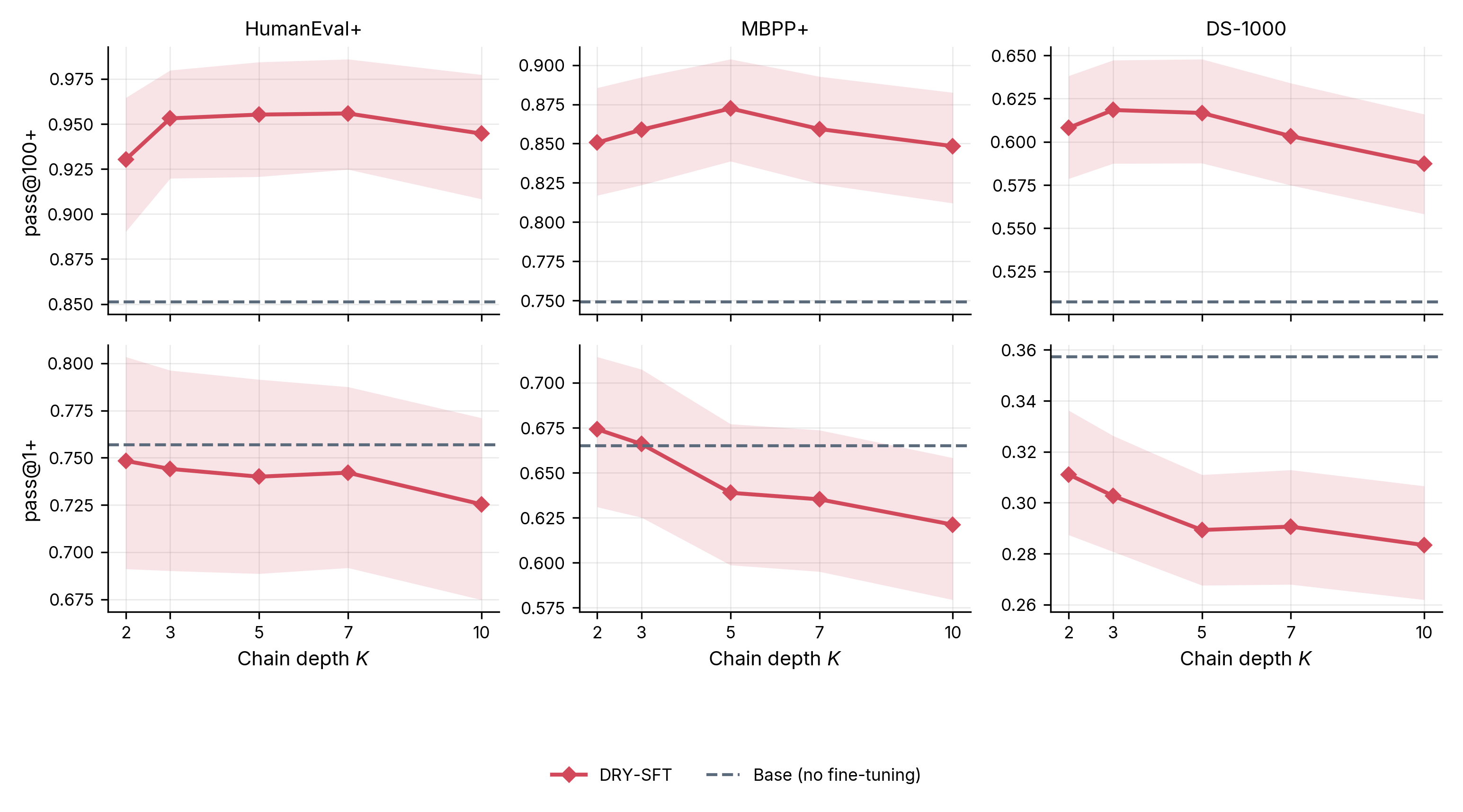}
  \caption{\passkplus{100} (\textbf{top}) and \passkplus{1} (\textbf{bottom}) for different $K$. Shaded regions are $95\%$ bootstrap confidence intervals over problems; dashed lines show the base model. We choose $K{=}5$ for fine-tuning. Appendix~\ref{app:kablation} reports the change compared to the base model.}
  \label{fig:kablation}
\end{figure}

\textbf{Shorter chains are not worse, and are usually better.} We sweep the chain depth $K$, holding every other choice fixed. Figure~\ref{fig:kablation} compares depths against $K{=}10$ on the same problem resample. Every depth raises \passkplus{100} over the base model by $\SweepGainMin$ to $\SweepGainMax$ points, and paired problem-bootstrap intervals exclude zero on all three benchmarks. $\SweepContrastShorter$ of the $\SweepContrastTotal$ contrasts favor the shorter chain, and \SweepContrastResolvedWord{} exclude zero: $K{=}3$, $K{=}5$, and $K{=}7$ on DS-1000 at $\SweepDSThreeDelta$ $[\SweepDSThreeLo,\SweepDSThreeHi]$, $\SweepDSFiveDelta$ $[\SweepDSFiveLo,\SweepDSFiveHi]$, and $\SweepDSSevenDelta$ $[\SweepDSSevenLo,\SweepDSSevenHi]$, and $K{=}5$ on MBPP+ at $\SweepMBFiveDelta$ $[\SweepMBFiveLo,\SweepMBFiveHi]$ points. No contrast favors $K{=}10$ by an interval excluding zero. Smaller $K$ retains fewer positions and targets, but each arm is a prefix of the same chain, so $K{=}10$ is the deepest arm rather than the reported one.

\textbf{The single-sample cost grows with depth.} \passkplus{1} declines overall as $K$ grows on all three benchmarks, although the point estimates are not strictly monotone. On the two EvalPlus benchmarks, the regression is not resolvable from zero at $K\le3$, and the coverage gain at short depths is close to free. On DS-1000, the regression is significant at every depth, but shrinks from $\DSSweepOneTen$ points at $K{=}10$ to $\DSSweepOneTwo$ points at $K{=}2$. The tradeoff is real on DS-1000, but only its size is tunable. Table~\ref{tab:kablation} gives every depth against the base on both metrics, with the intervals these statements are read from.

\begin{table}[ht]
  \centering
  \fontsize{10}{11}\selectfont
  \caption{Change in \passkplus{1} and \passkplus{100} compared to the base model for each $K$ with $95\%$ paired bootstrap confidence intervals over problems. $^{\dagger}$ indicates a statistically significant difference. The sweep holds training at three epochs rather than the validation-selected protocol of the reported runs (Appendix~\ref{app:epoch}), so its $K{=}5$ row differs from Section~\ref{subsec:main}.}
  \label{tab:kablation}
  \begin{tabular}{llrr}
  \toprule
  Benchmark & $K$ & $\Delta$\passkplus{1} & $\Delta$\passkplus{100} \\
  \midrule
  HumanEval+ & 2 & $-0.8$  $[-3.6, +2.0]$ & $+7.9$ $^{\dagger}$ $[+4.2, +12.2]$ \\
   & 3 & $-1.3$  $[-4.2, +1.6]$ & $+10.2$ $^{\dagger}$ $[+6.0, +14.8]$ \\
   & 5 & $-1.7$  $[-4.9, +1.6]$ & $+10.4$ $^{\dagger}$ $[+6.1, +15.2]$ \\
   & 7 & $-1.5$  $[-5.1, +2.2]$ & $+10.5$ $^{\dagger}$ $[+6.0, +15.1]$ \\
   & 10 & $-3.1$  $[-7.4, +1.1]$ & $+9.3$ $^{\dagger}$ $[+4.8, +14.3]$ \\
  \addlinespace
  MBPP+ & 2 & $+0.9$  $[-1.3, +3.3]$ & $+10.2$ $^{\dagger}$ $[+7.4, +13.3]$ \\
   & 3 & $+0.1$  $[-2.6, +2.6]$ & $+11.0$ $^{\dagger}$ $[+7.8, +14.5]$ \\
   & 5 & $-2.6$  $[-5.4, +0.2]$ & $+12.3$ $^{\dagger}$ $[+9.0, +15.9]$ \\
   & 7 & $-3.0$ $^{\dagger}$ $[-5.9, -0.0]$ & $+11.0$ $^{\dagger}$ $[+7.5, +14.7]$ \\
   & 10 & $-4.4$ $^{\dagger}$ $[-7.5, -1.3]$ & $+9.9$ $^{\dagger}$ $[+6.6, +13.6]$ \\
  \addlinespace
  DS-1000 & 2 & $-4.6$ $^{\dagger}$ $[-6.2, -3.2]$ & $+10.1$ $^{\dagger}$ $[+7.8, +12.3]$ \\
   & 3 & $-5.5$ $^{\dagger}$ $[-7.1, -3.9]$ & $+11.1$ $^{\dagger}$ $[+8.7, +13.4]$ \\
   & 5 & $-6.8$ $^{\dagger}$ $[-8.7, -5.0]$ & $+10.9$ $^{\dagger}$ $[+8.4, +13.5]$ \\
   & 7 & $-6.7$ $^{\dagger}$ $[-8.6, -4.9]$ & $+9.6$ $^{\dagger}$ $[+7.0, +12.1]$ \\
   & 10 & $-7.4$ $^{\dagger}$ $[-9.4, -5.5]$ & $+8.0$ $^{\dagger}$ $[+5.4, +10.5]$ \\
  \bottomrule
\end{tabular}

\end{table}

\textbf{The reported runs use $K{=}5$.} It sits at the top of the coverage plateau and well short of the depth where the cost to \passkplus{1} accelerates, whereas $K{=}10$ is the weakest setting on both axes: it gives up the most single-sample accuracy and never buys additional coverage. One confound remains. The sweep varies $K$, which varies the conditioning depth and the number of targets together, so the decline in \passkplus{1} is equally consistent with conditioning depth or with drift under more optimizer updates; on HumanEval+, the $K{=}10$ arm takes roughly $308$ optimizer updates, $154$ at $K{=}5$, and $62$ at $K{=}2$, and the matched i.i.d.\ SFT control is a depth-one arm. The sweep also holds training at three epochs, so longer chains are trained longer; the validation-selected protocol of the reported runs (Appendix~\ref{app:epoch}) removes this but does not retroactively disentangle the sweep. Separating the two would require varying the chain length at a fixed total number of targets, which we have not run. Target quality by chain position is in Appendix~\ref{app:positions}.

\section{Target quality along the chain}
\label{app:positions}

\begin{figure}[ht]
  \centering
  \includegraphics[width=0.55\textwidth]{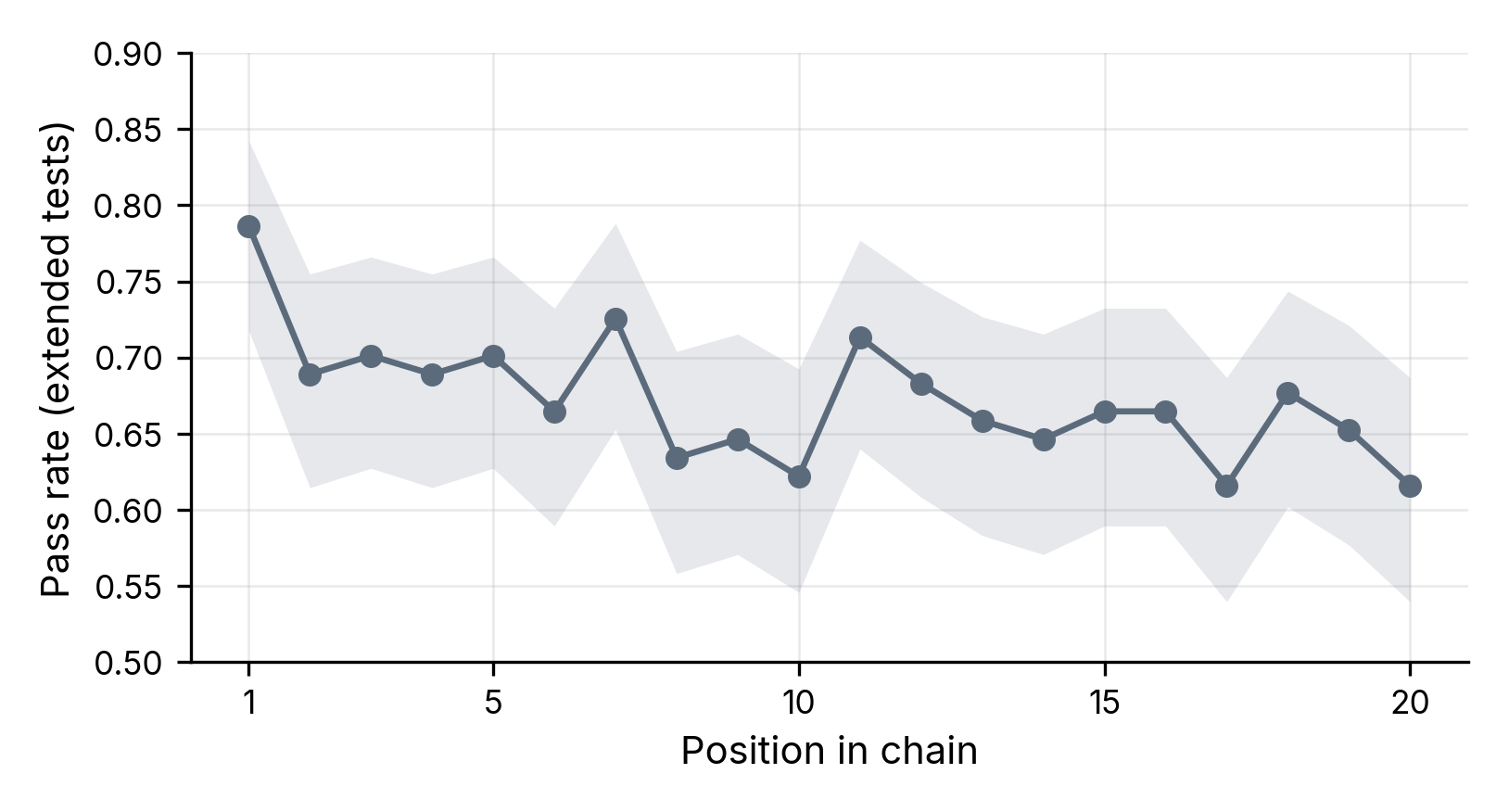}
  \caption{Pass rate at each position in a chain of $K{=}\PositionChainLength$ solutions, evaluated on the extended test suite. Shaded regions show $95\%$ confidence intervals.}
  \label{fig:positions}
\end{figure}

\textbf{Later positions pass less often, with no clean trend.} Figure~\ref{fig:positions} reports the pass rate at each position of a chain of length $K{=}\PositionChainLength$. It is $\PositionFirst$ at position 1 and $\PositionLast$ at position $\PositionChainLength$, with substantial non-monotonic variation in between. The lower endpoint is consistent with a tradeoff between differing from prior solutions and correctness, but it does not establish that the quality of later positions causes the $K{=}5$ optimum of Appendix~\ref{app:kablation}.

\FloatBarrier
\section{Correctness filtering before Stage~2}
\label{app:correctness-filtering}

We compare default \methodshort{} with correctness filtering after Stage~1 and before Stage~2. The unfiltered arm is the held-out \methodshort{} model of Section~\ref{subsec:transfer}. The filtered arm is retrained from Qwen3-4B-Instruct-2507 using the same $K{=}5$ chained candidates and seed-0 50/50 task splits. The filtered arm retains only targets passing both the original and extended tests on HumanEval+ and MBPP+, or the native tests on DS-1000. Filtering applies to training and validation targets, with validation task membership fixed before filtering. Stage~1 is unchanged: earlier failed attempts remain in the context for subsequent generations. The filtered arm uses the same training seed, optimizer settings, five-epoch budget, and minimum-validation-loss checkpoint selection as the held-out experiment. We evaluate on the untouched task half with $200$ independent samples per problem at $T{=}0.8$ and top-$p{=}0.95$.

\begin{figure}[ht]
  \centering
  \includegraphics[width=\linewidth]{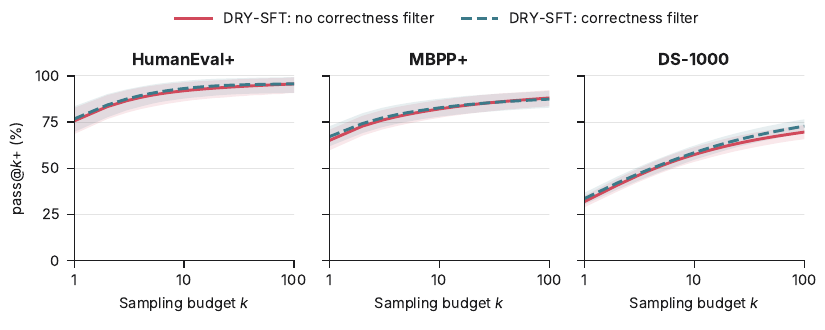}
  \caption{Held-out \passkplus{k} with and without correctness filtering before Stage~2. Both arms use the same generated targets before filtering and the same dataset split. Shaded regions are pointwise $95\%$ bootstrap confidence intervals over problems. Appendix~\ref{app:correctness-filtering} reports paired differences.}
  \label{fig:correctness-filtering}
\end{figure}

\textbf{Filtering changes coverage little.} Figure~\ref{fig:correctness-filtering} shows small differences between the two variants. At $k{=}100$, filtering changes coverage by $\HECorrectDeltaHundred$, $\MBCorrectDeltaHundred$, and $\DSCorrectDeltaHundred$ percentage points on HumanEval+, MBPP+, and DS-1000, respectively (Table~\ref{tab:correctness-filtering}). The paired intervals include zero on HumanEval+ and MBPP+, but exclude zero in favor of filtering on DS-1000. At $k{=}1$, filtering improves all three benchmarks with intervals excluding zero.

\begin{table}[ht]
  \centering
  \fontsize{10}{11}\selectfont
  \caption{Correctness-filtering ablation. Coverage is reported as a percentage; $\Delta$ is filtered minus unfiltered in percentage points, with pointwise $95\%$ paired problem-bootstrap intervals from $2{,}000$ resamples. Intervals quantify held-out task uncertainty for one training seed per arm. Values are rounded to one decimal place.}
  \label{tab:correctness-filtering}
  \begin{tabular}{llrrl}
    \toprule
    Benchmark & $k$ & Unfiltered & Filtered & $\Delta$ [$95\%$ CI] \\
    \midrule
    HumanEval+ & 1 & $\HECorrectUnfilteredOne$ & $\HECorrectFilteredOne$ & $\HECorrectDeltaOne$ [$\HECorrectLoOne$, $\HECorrectHiOne$] \\
               & 100 & $\HECorrectUnfilteredHundred$ & $\HECorrectFilteredHundred$ & $\HECorrectDeltaHundred$ [$\HECorrectLoHundred$, $\HECorrectHiHundred$] \\
    \addlinespace
    MBPP+ & 1 & $\MBCorrectUnfilteredOne$ & $\MBCorrectFilteredOne$ & $\MBCorrectDeltaOne$ [$\MBCorrectLoOne$, $\MBCorrectHiOne$] \\
          & 100 & $\MBCorrectUnfilteredHundred$ & $\MBCorrectFilteredHundred$ & $\MBCorrectDeltaHundred$ [$\MBCorrectLoHundred$, $\MBCorrectHiHundred$] \\
    \addlinespace
    DS-1000 & 1 & $\DSCorrectUnfilteredOne$ & $\DSCorrectFilteredOne$ & $\DSCorrectDeltaOne$ [$\DSCorrectLoOne$, $\DSCorrectHiOne$] \\
            & 100 & $\DSCorrectUnfilteredHundred$ & $\DSCorrectFilteredHundred$ & $\DSCorrectDeltaHundred$ [$\DSCorrectLoHundred$, $\DSCorrectHiHundred$] \\
    \bottomrule
  \end{tabular}
\end{table}

\textbf{The coverage gains need neither filtering nor failed targets.} Filtering retains $\HECorrectRetained/\HECorrectCandidates$, $\MBCorrectRetained/\MBCorrectCandidates$, and $\DSCorrectRetained/\DSCorrectCandidates$ candidate targets, including validation targets, on HumanEval+, MBPP+, and DS-1000. We do not replace discarded targets, so filtering also changes task coverage, optimizer-update counts, validation targets, and potentially the selected epoch. Both variants retain substantial coverage gains over the held-out base model (Section~\ref{subsec:transfer}). These gains require neither correctness filtering nor fitting test-failing Stage~2 targets. This comparison does not isolate the effect of incorrect targets at a fixed training budget, rule out a role for failed attempts in Stage~1 conditioning, or establish equivalent performance across training seeds.

\FloatBarrier
\section{Per-model coverage curves}
\label{app:second_model}

\begin{figure}[ht]
  \centering
  \includegraphics[width=\linewidth]{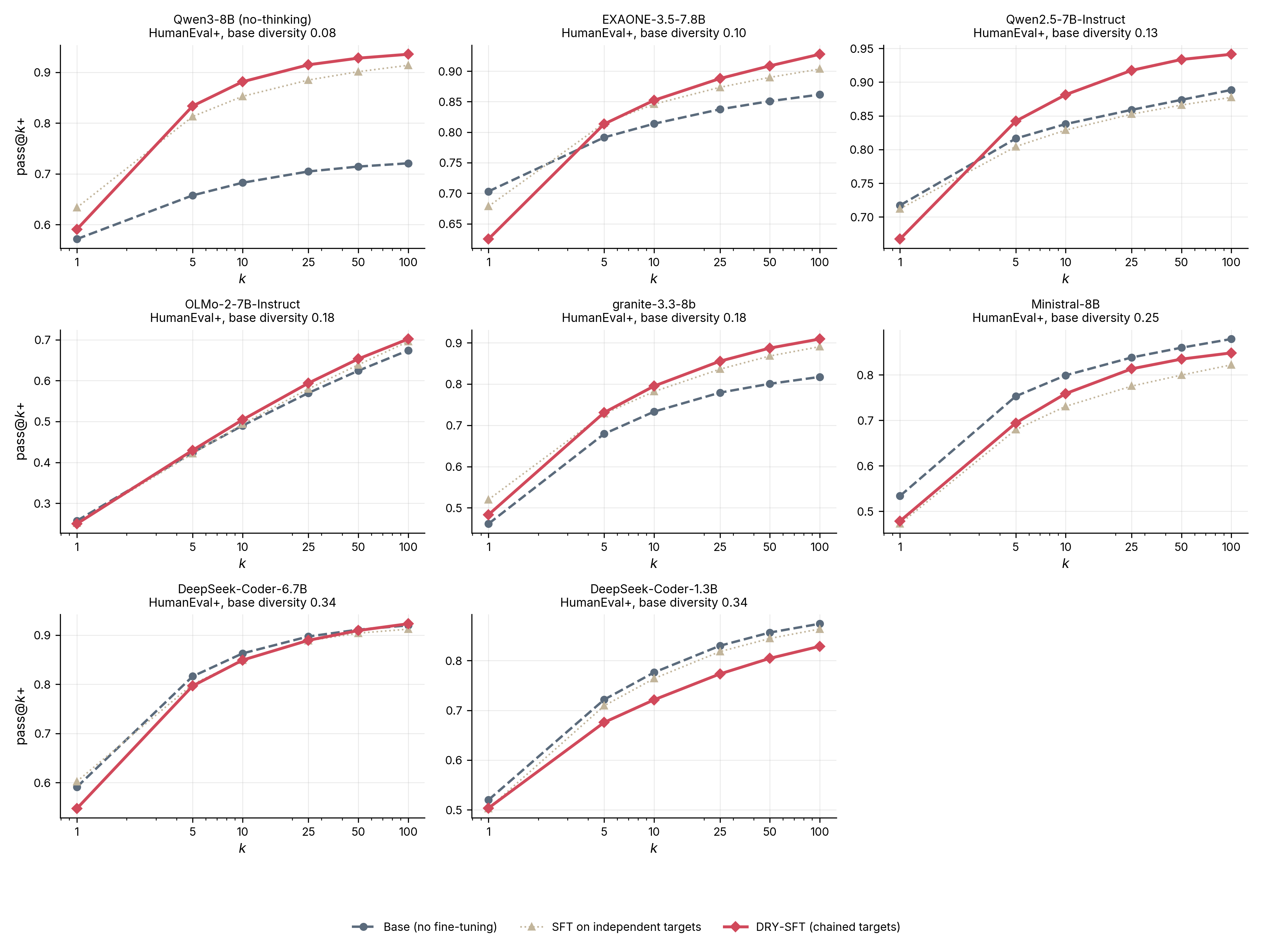}
  \caption{HumanEval+ \passkplus{k} for each model, comparing \methodshort{} with its base model and matched i.i.d.\ SFT control. Models with lower base structural diversity are shown first. Before evaluation, we remove output that is not code.}
  \label{fig:second_model}
\end{figure}

Section~\ref{subsec:scope} reports how a base model's structural diversity relates to the gain \methodshort{} produces across \ScopeFamiliesWord{} models. Figure~\ref{fig:second_model} gives the underlying \passkplus{k} curves, one per model, each against its own matched i.i.d.\ SFT control. The models are Qwen3-8B in its no-thinking mode \citep{qwen2025qwen3}, EXAONE-3.5-7.8B-Instruct \citep{lgai2024exaone35}, Qwen2.5-7B-Instruct \citep{qwen2024qwen25}, granite-3.3-8b-instruct \citep{granite2025granite33}, OLMo-2-7B-Instruct \citep{olmo2024olmo2}, Ministral-8B-Instruct-2410 \citep{mistral2024ministraux}, and deepseek-coder-6.7b-instruct and deepseek-coder-1.3b-instruct \citep{guo2024deepseekcoder}. Every model follows the same protocol as the primary model: $K{=}5$, $5$ unfiltered targets per problem, and no reward, verifier, or correctness filter at any point.

\FloatBarrier
\section{Prompt templates}
\label{app:prompts}

The exact prompts used in Stage~1 are different across benchmarks, as they need to match the output format of each benchmark. The instruction to differ from every prior solution is shared across benchmarks. We reproduce the templates below, as they appear in the generation code, before substituting placeholders. Each box holds one benchmark's system prompt above the divider and its user templates below, the first-position template being the one Stage~2 and evaluation reuse.

\begin{promptbox}{HumanEval+ Stage~1}
\begin{lstlisting}[style=promptstyle]
You are a Python coding assistant that generates diverse solutions. When shown previous solutions to a problem, you produce a new solution that uses a fundamentally different algorithm or approach.
\end{lstlisting}
\tcblower
\promptpart{FIRST POSITION}
\begin{lstlisting}[style=promptstyle]
Implement the following Python function.
Output ONLY the function body. No explanation, no markdown.

{prompt}
\end{lstlisting}
\medskip
\promptpart{LATER POSITIONS}
\begin{lstlisting}[style=promptstyle]
Here is a Python function to implement:

{prompt}

The following solutions have already been produced:

{previous_solutions}

Write a NEW solution that uses a DIFFERENT algorithm or approach from all of the above. Output ONLY the function body. No explanation, no markdown.
\end{lstlisting}
\end{promptbox}
For HumanEval+, we serialize each prior target as \verb|--- Solution {i} ---| followed by the complete function prompt and generated body.

\begin{promptbox}{MBPP+ Stage~1}
\begin{lstlisting}[style=promptstyle]
You are a Python coding assistant that generates diverse function implementations. When shown previous solutions to a problem, produce a new solution that uses a different approach. Output ONLY full Python function code, no markdown.
\end{lstlisting}
\tcblower
\promptpart{FIRST POSITION}
\begin{lstlisting}[style=promptstyle]
Solve the following MBPP task.
Output ONLY the full Python function definition needed to solve it.

{prompt}
\end{lstlisting}
\medskip
\promptpart{LATER POSITIONS}
\begin{lstlisting}[style=promptstyle]
Solve the following MBPP task:

{prompt}

The following solutions have already been produced:

{previous_solutions}

Write a NEW solution that uses a DIFFERENT algorithm or approach from all above. Output ONLY full Python function code. No markdown.
\end{lstlisting}
\end{promptbox}
For MBPP+, we serialize each prior target as \verb|--- Solution {i} ---| followed by the generated full function.

\begin{promptbox}{DS-1000 Stage~1}
\begin{lstlisting}[style=promptstyle]
You are a data-science coding assistant that generates diverse completions. When shown previous completions, produce a new completion that uses a fundamentally different approach. Output ONLY the completion code in a ```python``` block.
\end{lstlisting}
\tcblower
\promptpart{FIRST POSITION}
\begin{lstlisting}[style=promptstyle]
Complete the following data-science coding problem.
Output ONLY the missing code completion.

{prompt}
\end{lstlisting}
\medskip
\promptpart{LATER POSITIONS}
\begin{lstlisting}[style=promptstyle]
Complete the following data-science coding problem:

{prompt}

The following completions have already been produced:

{previous_solutions}

Write a NEW completion that uses a DIFFERENT approach from all above. Output ONLY the missing code completion.
\end{lstlisting}
\end{promptbox}
For DS-1000, each prior target is serialized as a numbered solution with a fenced Python code block.

\paragraph{Stage~2 and evaluation.} For every example in Stage~2, we pair the generated target with the bare benchmark problem and a neutral correctness-format system prompt, without any prior solution or diversity prompt. The user side is the first-position template of the matching benchmark above; only the system prompt changes, and MBPP+ is the one benchmark whose evaluation prompt differs from the one it was trained under.

\begin{promptbox}[unbreakable]{HumanEval+ Stage~2 and evaluation}
\begin{lstlisting}[style=promptstyle]
Complete the following Python function. Output ONLY the function body (the code that goes inside the function). Do not repeat the function signature or docstring. Do not include any explanation or markdown formatting.
\end{lstlisting}
\end{promptbox}

\begin{promptbox}[unbreakable]{MBPP+ Stage~2}
\begin{lstlisting}[style=promptstyle]
Write a correct Python function implementation. Output ONLY the full function definition. No explanation and no markdown.
\end{lstlisting}
\end{promptbox}

\begin{promptbox}[unbreakable]{MBPP+ evaluation}
\begin{lstlisting}[style=promptstyle]
Write a correct Python function implementation. Output ONLY valid Python code for the full function definition. No explanation and no markdown.
\end{lstlisting}
\end{promptbox}

\begin{promptbox}[unbreakable]{DS-1000 Stage~2 and evaluation}
\begin{lstlisting}[style=promptstyle]
You are an expert data scientist writing Python code. Complete the unfinished code. Output ONLY the code that should replace the incomplete portion, preferably in a ```python``` block. Do not repeat the provided context.
\end{lstlisting}
\end{promptbox}

\end{document}